\documentclass[runningheads]{llncs}

\usepackage{eccv}

\usepackage{eccvabbrv}

\usepackage{amsmath,amsfonts,bm}

\usepackage{graphicx}
\usepackage{booktabs}
\usepackage{makecell}
\usepackage{pifont}
\usepackage{wrapfig}

\usepackage[accsupp]{axessibility}

\usepackage{hyperref}

\usepackage{orcidlink}

\newcommand{\cmark}{\ding{51}}%
\newcommand{\xmark}{\ding{55}}%

\newcommand{\R}{\mathbb{R}}

\newcommand{\citep}[1]{\cite{#1}}
\newcommand{\citet}[1]{\cite{#1}}

\newcommand{\FH}[1]{}
\newcommand{\WS}[1]{}

\begin{document}

\title{Real-VAS: a Real-World Video Amodal Segmentation Dataset}

\titlerunning{Real-VAS: Real-World Video Amodal Segmentation}

\author{Finlay G. C. Hudson\inst{1} \and
William A. P. Smith\inst{1}}

\authorrunning{F.G.C.~Hudson and W.A.P.~Smith}

\institute{Department of Computer Science, University of York, York, United Kingdom\\
\email{\{fgch500,william.smith\}@york.ac.uk}}

\maketitle

\begin{abstract}
Amodal video object segmentation is fundamentally limited by the absence of datasets that combine real-world complexity with precise ground-truth annotations. To address this, we present Real Video Amodal Segmentation (Real-VAS), a new large-scale, zero-shot evaluation dataset built from composited real-image video. We introduce a novel data generation pipeline that enables the creation of \emph{programmatically consistent} amodal ground truth for real-world footage, in which the object mask, occlusion boundary, and composited image are all derived from a single unoccluded source, without relying on human estimation or expensive 3D reconstruction. Our dataset is structured into two challenging scenarios: dynamic Occlusion, created by compositing two clips of moving objects, and a unique Container category featuring complex, physically constrained interactions. These container scenarios--On Surface Containment, Articulated Containment, and Mobile Containment--allow us to generate precise ground truth by simulating an object's motion based on its container's tracked transformation. As a result, Real-VAS provides a diverse and challenging benchmark for evaluating amodal segmentation models on realistic video with the precision of synthetic data. The dataset and generation code are publicly available at \url{https://finlay-hudson.github.io/real_vas}.
\end{abstract}
\section{Introduction}
\label{sec:intro}

Video Object Segmentation (VOS) is the problem of densely tracking the visible pixels of an object. Methods such as SAM2~\cite{ravi2024sam} and Mask2Former~\cite{cheng2022masked} have made remarkable progress on this problem in recent years, offering very high accuracy and robustness. However, in humans our visual system has a strong notion of \textit{object permanence}~\cite{an2022object} -- the idea that objects maintain their identity and continuity over time, regardless of visibility -- allowing us to predict where an object exists and how it may have deformed, even when occluded. This concept is called \textit{amodal completion}~\cite{ao2023image, nanay2018importance} and is the focus of this paper. Robust amodal completion is critical for many real-world applications, from autonomous driving, where vehicles must track occluded pedestrians, to robotic manipulation, where a robot needs to infer an object's full shape to grasp it in a cluttered environment.

Progress on machine Video Amodal Segmentation (VAS), \ie densely tracking object segmentations \emph{through} occlusions, has received relatively less attention and progress has not been as rapid~\cite{zhu2019robustness, kortylewski2020compositional}. One of the reasons for this limited progress is the lack of sufficiently challenging datasets with ground truth segmentations behind occlusions. Existing datasets suffer from several important weaknesses. On the one hand, synthetic datasets provide exact ground truth but are limited in realism and diversity, particularly for dynamic objects. On the other hand, obtaining ground truth segmentations for objects behind occlusions in real-world video is an extremely difficult and resource intensive task~\cite{vanhoorick2023tcow}. Possible solutions include multi-camera or 3D setups to record ground truth, which is not scalable, or human best-guess annotations behind occlusions, which is not accurate. The real-world benchmarks that are available do not provide accurate (or any) ground truth segmentation behind occlusions. Instead, they either provide ground truth only on unoccluded frames (in which case the benchmark is really testing reidentification~\cite{zheng2016mars, bergmann2019tracking, zhang2022bytetrack, zhang2021fairmot} rather than amodal segmentation) or rely on hand-drawn, approximate masks. Finally, existing datasets do not contain sufficiently challenging scenarios such as occlusion by a container (which requires reasoning about the motion of the container) or objects never re-appearing after occlusion (which requires extrapolation without relying on reidentification).

\begin{figure}[!t]
    \centering
\includegraphics[width=\textwidth,trim=124px 121px 122px 118px,clip=true]{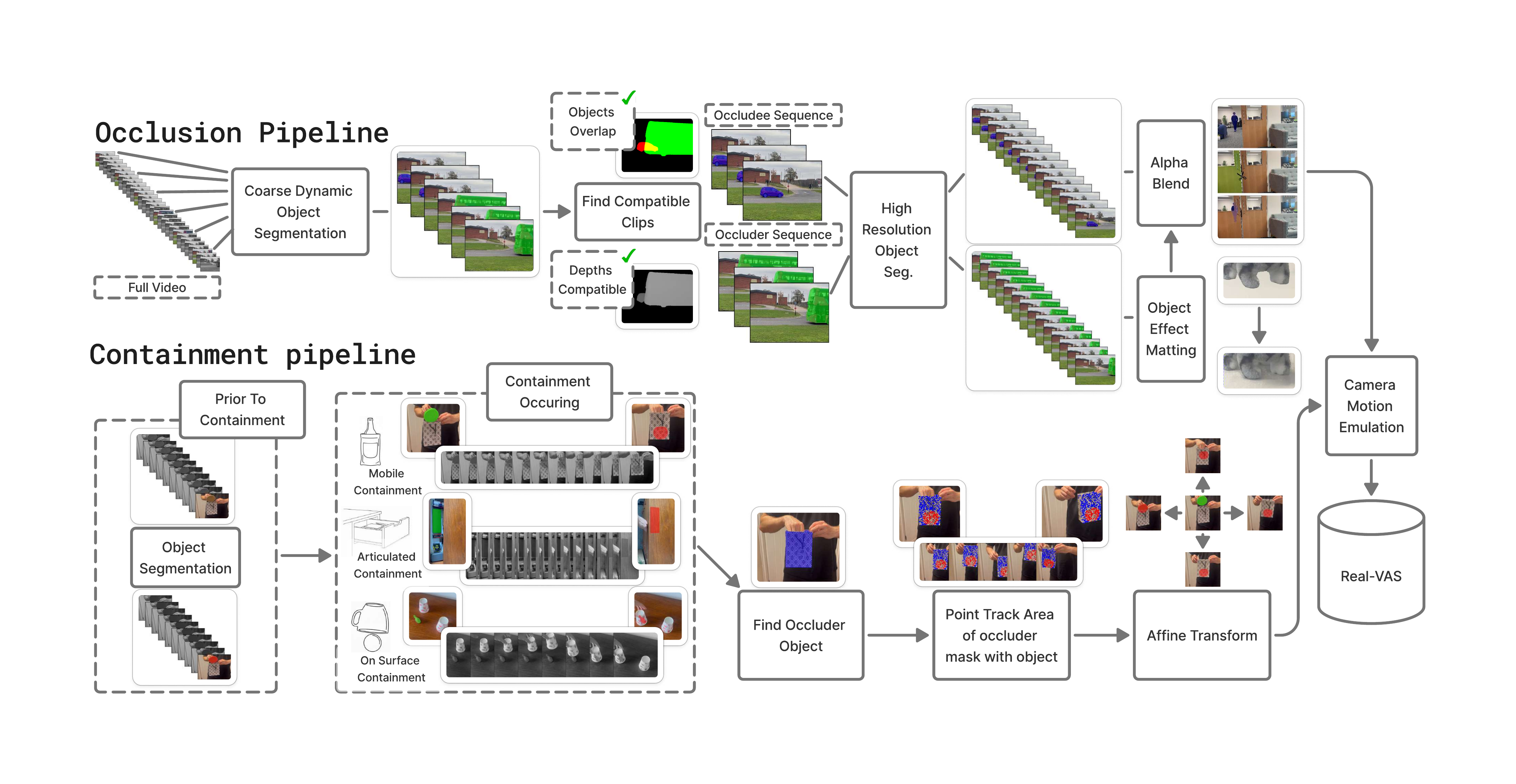}
    \caption{An overview of our Real-VAS data generation pipeline. Occlusion pipeline: from static camera videos, we automatically find compatible pairs of clips that can be composited to form realistic occlusions. Occluder objects and their effect (\eg shadows) are composited over occludee objects for which accurate ground truth is computed from the unoccluded sequence. Containment pipeline: we identify three scenarios in which tracking the container is sufficient to provide ground truth tracking of the contained object. For both pipelines, additional camera motion can be subsequently emulated by cropping to a narrower moving field of view.}
    \label{fig:real-vas-fig}
\end{figure}

In this paper we propose scalable methods for creating a large-scale VAS dataset using real-world videos and without specialist acquisition setups. While synthetic datasets naturally offer a greater sheer volume of scenes, Real-VAS bridges a critical sim2real gap by focusing on the density and severity of real-world occlusions. Specifically, Real-VAS provides over 10,600 occluded frames, more than three times the number of occluded frames found in the next largest real-world competitor (see \cref{tab:dataset_comparison}). With an occlusion prevalence of 72.53\% and a mean occlusion severity of 67.33\%, it offers a substantially harder and more comprehensive benchmark than existing datasets. We propose two pipelines (see \cref{fig:real-vas-fig}). The first allows arbitrary rigid or non-rigid, possibly dynamic occluders to be photorealistically composited over arbitrary occludee objects. The second allows complex occludee-container relationships to be accurately tracked. We use these methods to create the most challenging VAS dataset to date, Real-VAS,
which is publicly available along with code for dataset generation.\footnote{Dataset: \url{https://huggingface.co/datasets/fhudson96/Real-VAS}; code: \url{https://github.com/finlay-hudson/Real-VAS}}

\section{Related Work}
\label{sec:related_work}

\begin{table}[b]
\centering
\resizebox{\textwidth}{!}{%
\begin{tabular}{@{}lccccccccc@{}}
\toprule
& \multicolumn{4}{c}{\textbf{Real Datasets}} & \phantom{a} & \multicolumn{4}{c}{\textbf{Synthetic Datasets}} \\
\cmidrule(r){2-5} \cmidrule(l){7-10}
 & \textbf{Real-VAS} & \textbf{COCOA} & \textbf{COCOA-cls} & \textbf{D2S} & & \textbf{MOVi-MC-AC} & \textbf{SAIL-VOS 3D} & \textbf{SAIL-VOS} & \textbf{DYCE} \\
\textbf{Metric} & \textit{Ours} & \cite{zhu2017semantic} & \cite{follmann2019learning} & \cite{follmann2018mvtec} && \cite{moore2025training} & \cite{HuCVPR2021} & \cite{hu2019sail} & \cite{ehsani2018segan} \\
\midrule
Image or Video & Video & Image & Image & Image && Video & Video & Video & Image \\
Accurate Annotations & \cmark & \xmark & \xmark & \xmark && \cmark & \cmark & \cmark & \cmark \\
Scene Images &  14,677 &  5,073 & 3,499 & 5,600 && 293,904 & 237,611 & 111,654 & 5,500 \\
Unique Objects & 160 & - & 80 & 60 && 1,033 & 178 & 162 & 79 \\
Occlusion Prevalence & 72.53\% & 60.7\% & 49.0\% & 56.9\% && 69.3\% & - & 87.2\% & 82.3\% \\
Mean Occlusion Severity & 67.33\% & 18.8\% & 10.7\% & 15.0\% && 45.2\% & - & 56.3\% & 27.7\% \\
\bottomrule
\end{tabular}%
}
\caption{Comparison of amodal segmentation datasets. Occlusion Prevalence measures how often each instance is occluded. Mean Occlusion Severity measures what proportion of an object is occluded on average when they are occluded.}
\label{tab:dataset_comparison}
\vspace{-0.7cm}
\end{table}

\noindent\textbf{Amodal Segmentation Methods.}
Current amodal segmentation methods can be categorized by architectural paradigm and temporal context usage. Architecturally, feedforward models like TCOW~\cite{vanhoorick2023tcow} directly predict amodal states, while generative approaches treat completion as conditional in-painting using diffusion priors. Image-based diffusion methods~\cite{zhan2020self, ozguroglu2024pix2gestalt, zhan2024amodal} process frames independently, whereas video-based approaches~\cite{hudson2024track, chen2025using, lee2025generative} operate offline on sequences for temporal consistency. These methods rely on various contextual cues, often requiring additional inputs like modal masks~\cite{ozguroglu2024pix2gestalt, zhan2024amodal, hudson2024track, chen2025using, lee2025generative} or bounding boxes~\cite{zhan2024amodal}, and employ techniques like depth estimation~\cite{ozguroglu2024pix2gestalt, hudson2024track, chen2025using} or target region estimation~\cite{hudson2024track}.

Taking an alternative approach, ~\cite{yao2022self} introduced a self-supervised framework (SaVos) for amodal video object segmentation. By exploiting the observation that occluded parts of an object in one frame are often visible in other frames, SaVos leverages motion cues and optical flow to learn amodal completions without requiring manually annotated amodal ground truth, demonstrating strong results on both synthetic and real-world benchmarks.

The field has recently seen significant advances, particularly in adapting foundation models. SAMBA~\cite{liu2025samba} introduces the first SAM-based amodal segmentation foundation model for zero-shot inference, while SAMEO~\cite{tai2025sameo} adapts SAM as a versatile amodal mask decoder, also introducing a large-scale synthetic training dataset. For video amodal completion, TACO~\cite{lu2025taco} repurposes video diffusion models for in-the-wild scenarios, and A2VIS~\cite{tran2025a2vis} incorporates amodal representations into video instance segmentation. Other work has extended amodal reasoning to language-guided settings~\cite{li2025aura, ao2025openworldamodal} and shown that amodal segmentation emerges when scaling diffusion models for perceptual tasks~\cite{ravishankar2025scaling}.

Among the state-of-the-art methods evaluated in our benchmarks, Diffusion VAS~\cite{chen2025using} and TABE~\cite{hudson2024track} formulate amodal completion as a conditional video generation task. Diffusion VAS leverages pre-trained diffusion priors combined with depth estimation to hallucinate occluded regions while maintaining temporal consistency. Similarly, TABE employs an offline video diffusion architecture, leveraging global temporal context to track object states through severe and complex occlusions.

\noindent\textbf{Amodal Segmentation Datasets.}
Existing datasets (\cref{tab:dataset_comparison}) face a trade-off between annotation accuracy and real-world fidelity. Synthetic datasets achieve perfect ground truth via physics modeling~\cite{vanhoorick2023tcow, hu2019sail, reddy2022walt, fan2023rethinking, mohan2022amodal}, motion tools~\cite{tangemann2021unsupervised, yao2022self, gao2023coarse, girdhar2019cater, shamsian2020learning}, and image occlusion~\cite{follmann2018mvtec, ehsani2018segan}, but suffer from sim2real gaps~\cite{hofer2021sim2real, wechsler2024bridging}.
Real-world approaches face annotation challenges since occluded regions require human estimation rather than ground truth measurement---a process that is time-intensive, inaccurate, and subjective~\cite{zhu2017semantic, qi2019amodal, follmann2019learning}. Large-scale tracking benchmarks like MOTChallenge \cite{dendorfer2021motchallenge} and TAO-Amodal \cite{hsieh2023tao} rely on human estimates of bounding boxes rather than dense, pixel-level amodal masks. 3D acquisition methods~\cite{zhan2024amodal, li2023muva} provide precise masks but lack scalability. A compositing approach has also been adopted for the related task of amodal depth estimation~\cite{li2025amodaldepthanything}, though not for pixel-level amodal segmentation in video.

\begin{wraptable}{r}{0.5\textwidth}
\centering
\vspace{-0.4cm}
\begin{tabular}{@{}cccc@{}}
\toprule
     & Rubric & Rubric & Rubric  \\
     & Office & Cup Games & DAV/YTB \\ \midrule
TCOW & 69.4          & 38.3             & 52.8           \\
SAM2 & \textbf{72.7} & \textbf{55.8}    & \textbf{68.9}  \\ \bottomrule
\end{tabular}
\caption{Comparison of TCOW~\cite{vanhoorick2023tcow} and SAM2~\cite{ravi2024sam} mean IoU on amodal segmentation datasets. Despite SAM2 providing only \emph{modal} segmentation, it outperforms TCOW's amodal result.}
\label{tab:sam2_beats_tcow}
\vspace{-0.2cm}
\end{wraptable}

Compositing approaches, successful in other vision tasks~\cite{walawalkar2020attentive, devries2017improved, bochkovskiy2020yolov4, zhao2020differentiable, kini2024ct}, often rely on simplistic overlays in amodal segmentation~\cite{ozguroglu2024pix2gestalt, lee2025generative}, creating realism gaps. Additionally, existing datasets focus on simple ``pass-by'' occlusions, neglecting challenging containment scenarios where objects physically interact~\cite{vanhoorick2023tcow}. TCOW's real-world evaluation only annotates unoccluded frames, reducing the task to re-identification rather than true amodal segmentation (see \cref{tab:sam2_beats_tcow}).

To address these significant gaps in existing work, we introduce Real-VAS, a novel dataset and data generation pipeline for zero-shot video amodal segmentation. Crucially, Real-VAS operates at a scale previously unseen in real-world amodal benchmarks, containing over 3 times more occluded frames than its closest competitor. Our approach is designed to synthesise the strengths of prior methods while mitigating their respective drawbacks. Our \textit{Occlusion} methodology uses a compositing pipeline on real-world footage, achieving the precise ground truth of synthetic data without the sim2real gap or the imprecision and high cost of manual annotation. To overcome the realism gap of prior compositing work, our Occlusion subset features dynamic interactions between moving objects and includes emulated camera motion. Furthermore, to move beyond simple pass-by events and benchmark physical reasoning, we introduce the novel \textit{Containment} category with three physically-grounded scenarios---On Surface Containment, Articulated Containment, and Mobile Containment.

\section{Task}
\label{sec:task}

The \emph{zero-shot video amodal segmentation} (ZS-VAS) task is defined as follows. We are given as input a $T$-frame video $\mathbf{x}\in\R^{T\times H\times W\times 3}$ along with a binary query mask $\mathbf{m}_\text{q}\in\{0,1\}^{H\times W}$ which segments the object of interest in some reference frame (assumed to be frame 1). It is assumed that the object is completely unoccluded in the reference frame. The query mask may itself have been computed from a prompt of some form (\eg text or point clicks). The goal is to learn a function $f$ that estimates amodal segmentation masks for the tracked object in every frame,
$\mathbf{m}_\text{o} = f(\mathbf{x},\mathbf{m}_\text{q})$,
where $\mathbf{m}_\text{o}\in\{0,1\}^{T\times H\times W}$ is the binary object mask in every frame. This must delineate the object even behind occluders and even when none of the object is visible based on information from surrounding frames and the context of the object itself. This initialisation differs from datasets like TAO-Amodal (amodal object tracking) and SAIL-VOS (synthetic amodal object segmentation), where the object of interest may be occluded from the start, often requiring a model to rely on pre-existing class knowledge for identification. The zero-shot paradigm is therefore class-agnostic; it evaluates the more general and realistic ability of a model to segment any novel object based purely on the provided query, rather than its familiarity with specific object categories.

We created this dataset to be an evaluation task, providing a wide-ranging benchmark of real-world scenarios, dynamic object classes, and physically complex interactions designed to test the limits of zero-shot amodal segmentation and aid in the development of the field.

\subsection{Evaluation Metrics}

\begin{wraptable}{r}{0.5\textwidth}
\centering
\vspace{-0.4cm}
\begin{tabular}{@{}lccc@{}}
    \toprule
     & SAM2 & TABE & Diffusion VAS \\
    \midrule
mIoU & 0.621 & 0.542 & 0.655 \\
\hline
$\text{mIoU}_\text{occ}$ & 0.066 & 0.207 & 0.262 \\
\bottomrule
\end{tabular}
\caption{Standard mIoU is a misleading metric for amodal segmentation. A purely modal method (SAM2) achieves a high overall score while failing on occluded regions ($\text{mIoU}_\text{occ}$).}
\label{tab:why_not_miou}
\vspace{-0.2cm}
\end{wraptable}

The standard metrics used for evaluating amodal segmentation are mean Intersection-over-Union (mIoU) and mIoU computed over only occluded pixels, referred to as $\text{mIoU}_\text{occ}$~\cite{chen2025using, gao2023coarse, fan2023rethinking}. However, pooled mIoU is misleading in this regime: high scores on unoccluded frames obscure poor performance on heavily occluded ones, allowing a purely modal tracker like SAM2 to score deceptively well (0.621 in \cref{tab:why_not_miou}) while completely failing under total occlusion ($\text{mIoU}_\text{occ}=0.066$). For this reason we report a hierarchy of three metrics, each targeting a different failure mode. $\text{mIoU}_\text{fo}$ is our \emph{primary} metric: it measures overall tracking stability over all frames with any level of occlusion, including partial occlusion. $\text{mIoU}_\text{ffo}$ is a stricter variant restricted to \emph{fully occluded} frames, isolating the model's ability to hallucinate the object during total invisibility. $\text{mIoU}_\text{occ}$ measures per-pixel precision on the occluded region only, and so is not inflated by correctly predicting visible pixels:
\[
\text{mIoU}_\text{fo} = \tfrac{\sum_{i \in F_\text{occ}} |\hat{A}_i \cap A_i|}{\sum_{i \in F_\text{occ}} |\hat{A}_i \cup A_i|},
\quad
\text{mIoU}_\text{ffo} = \tfrac{\sum_{i \in F_\text{full}} |\hat{A}_i \cap A_i|}{\sum_{i \in F_\text{full}} |\hat{A}_i \cup A_i|},
\]\[
\text{mIoU}_\text{occ} = \tfrac{\sum_{i \in F_\text{occ}} |(\hat{A}_i - M_i) \cap (A_i - M_i)|}{\sum_{i \in F_\text{occ}} |(\hat{A}_i - M_i) \cup (A_i - M_i)|},
\]
where $M_i$ is the ground-truth \textbf{modal mask}, $\hat{A}_i$ the \textbf{predicted amodal mask}, and $A_i$ the \textbf{ground-truth amodal mask} for frame $i$. We define $F_{occ}$ as the set of \textbf{occluded frames} and $F_\text{full} \subseteq F_\text{occ}$ as the set of \textbf{fully occluded frames} (for which $M_i = \emptyset$).
\section{Real-VAS Dataset}
\label{sec:dataset}

\begin{figure}[!t]
    \centering
\includegraphics[width=\textwidth]{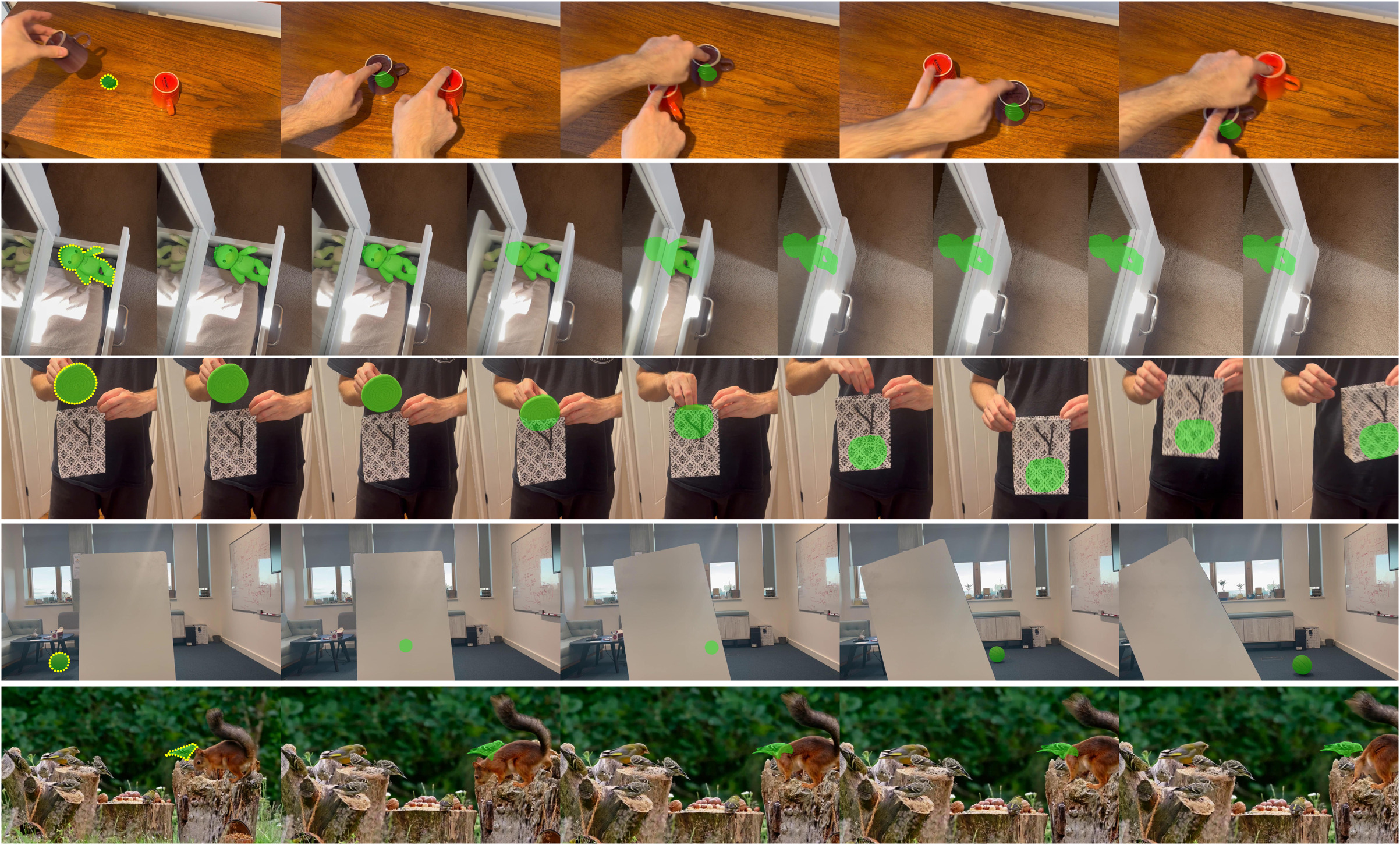}
    \caption{Qualitative examples from the Real-VAS dataset. The yellow dashed line indicates the initial query mask, followed by the ground-truth amodal segmentation frames. The rows showcase our scenarios: On Surface Containment, Articulated Containment, Mobile Containment, and two Dynamic Occlusion examples, one captured in-house (row 4) and one sourced from the web (row 5).}
    \label{fig:dataset_examples}
\vspace{-0.7cm}
\end{figure}

We introduce Real-VAS, a novel dataset with precise ground-truth annotations for the evaluation of video amodal segmentation. As shown in \cref{fig:real-vas-fig}, Real-VAS is constructed using two distinct pipelines, `occlusion' and `container', to test performance across a diverse range of amodal scenarios, while remaining automatic and scalable to enable the creation of large-scale datasets. To ensure the final quality of the dataset, we include two manual quality control checks (Appendix~\ref{sec:qc}).

We use two sources of raw data: controlled scenes captured in-house and scraping diverse videos from public video repositories. This hybrid approach allows us to combine staged, highly controlled interaction scenarios with the vast diversity of real-world footage. Our in-house data was captured using a tripod-mounted iPhone 15, recording at a high frame rate (slow-motion) with locked focus and exposure to ensure consistent framing and minimal motion blur for dynamic objects. For the externally sourced data, we utilised footage from both existing large-scale video datasets~\cite{personpath22, chavdarova2017wildtrack} and in-the-wild videos from public platforms such as YouTube, filtered to provide videos from static cameras.

\subsection{Real-VAS Occlusion}
We use a compositing strategy to obtain precise masks for occluded objects. For scenes we capture ourselves, we record two clips with a static camera: one showing the object to be occluded (the ``occludee'') without obstruction, and a second clip showing the ``occluder''. For scraped video data, we propose an automated pipeline for finding suitable pairs of sub-clips that contain a compatible occluder and occludee. We begin by feeding text prompts from a predefined list of dynamic object classes (see Appendix~\ref{sec:classlist}) into Language Segment-Anything~\cite{Ledeiros_Lang-Segment-Anything}. This open-vocabulary model segments the corresponding objects by combining a grounding detector~\cite{zhang2023grounding} with SAM2. The resulting masks are then tracked across all frames with SAM2.

To ensure the dataset is both high-quality and challenging, a dynamic object checker filters for sequences where both the occluder and occludee are dynamic. To determine if an object is dynamic, we calculate a mean motion score $\mathcal{S}$ based on the trajectories of its points over time, gathered using CoTracker3~\cite{karaev2024cotracker3}.
Let the set of trajectories for a single object be represented by $\mathbf{T} \in \mathbb{R}^{T \times P \times 2}$, where $T$ is the total number of frames (time steps), $P$ is the number of points tracked on the object, $\mathbf{p}_{t,i} \in \mathbb{R}^{2}$ denotes the $(x, y)$ coordinates of the $i$-th point at frame $t$. We define the \emph{mean motion score}, $\mathcal{S}$, as the average displacement (Euclidean distance) of all points over all frames:
$$
\mathcal{S} = \frac{1}{PT} \sum_{i=1}^{P} \sum_{t=1}^{T-1} \left\| \mathbf{p}_{t+1, i} - \mathbf{p}_{t, i} \right\|_2.
$$
An object is classified as \textbf{dynamic} if its mean motion score exceeds a predefined minimum motion threshold,
$
\mathcal{S} > \tau
$.
This formulation effectively captures the object's overall movement by averaging the speed of its constituent points across the entire video sequence.

Now that we have segmentation masks for dynamic objects in the video we can find where composited occlusion can be created. The core purpose of this method is to \textbf{discover and validate high-quality object occlusion events} by finding the best way to temporally align two separate video sequences of object masks ($M_1$ and $M_2$). The process is structured into two main stages: Candidate Generation and Rigorous Filtering.

\subsubsection{Candidate Generation via Cross-Correlation.}

The algorithm first identifies the most promising temporal alignments by calculating an \textbf{overlap score} for every possible time shift ($\Delta t$) between the two valid mask sequences, \ie a \textbf{cross-correlation}. The score $S(\Delta t)$ for a given shift is the sum of the spatial intersection (the overlapping pixel area) over the duration of the sequence:
\begin{equation*}
    S(\Delta t) = \sum_{t} | M_1[t] \cap M_2[t - \Delta t] |,
\end{equation*}
where $| \cdot |$ denotes the number of pixels in the intersection. We compute this score for all possible shifts and sort them, creating a ranked list of the best potential occlusion events.

\subsubsection{Rigorous Filtering Pipeline.}

Next, the algorithm iterates through the top-ranked candidates and subjects each one to a comprehensive set of validation checks to ensure the resulting composite scene is physically plausible, interesting, and non-redundant.
For each candidate, the objects are first assigned \textbf{foreground (FG)} and \textbf{background (BG)} roles based on their relative depth values. The sequence is then trimmed or rejected if it fails any of the following key tests:
\noindent\textbf{Physical plausibility:} We perform several automated checks using monocular depth estimates from Depth Anything V2~\cite{yang2024depth}. First, we prevent \textit{Reverse Occlusion} by analysing the pixel region where the foreground (FG) and background (BG) masks overlap. The sequence is rejected if any FG pixel in this region has a greater depth than its corresponding BG pixel. Second, our \textit{``On Top'' Detection} filters out scenarios where objects appear to improperly occupy the same 3D space. Finally, we check for \textit{External Occlusion} by detecting if an object's size unexpectedly reduces. If the depth indicates a closer surface at the point of shrinkage, we infer an external occlusion and discard the candidate.

\noindent\textbf{Quality \& Interest Checks:} First, every sequence must have a clean start, meaning the background object is fully visible in the initial frame. Next, the interaction must result in sufficient occlusion, where the occluded area exceeds a minimum threshold.

\noindent\textbf{Redundancy Checks:} A similarity test discards candidate sequences that are too visually similar to previously accepted ones, while a frame overlap check prevents the same source frames from being used for both the foreground and background objects.

Only candidates that successfully pass this entire pipeline are considered valid and passed forward to the next stage of the pipeline.

\noindent\textbf{High Resolution Sequence Generation.} To ensure computational efficiency, the preceding steps operated on video clips temporally downsampled to 1--3\,fps. This frame rate provided sufficient temporal resolution to find candidate overlaps while minimising processing time. To convert the coarse, low-framerate mask sequences (1--3\,fps) into high-quality masks at the video's native framerate, we employ a propagation-based approach once again using SAM2. This process involves an initialisation step followed by a focused, fine-grained segmentation pass. The process begins by identifying the first valid mask in the low-framerate sequence to serve as an initial prompt. However, temporal downsampling can miss the true first appearance of an object. To correct this, we perform a preliminary backwards process: the segmentation model is run on a reversed video at high framerate, immediately preceding this initial mask. This refinement step pinpoints the object's true starting frame and provides a more accurate initial mask for the main forward process. To generate fine-grained masks with high efficiency and precision, we first create a dense sequence of guiding bounding boxes. This is done by linearly interpolating the sparse set of boxes from the low-framerate sequence, one for every high-framerate frame, and adding a motion-compensation padding. The segmentation model is then applied exclusively within these tight, content-focused crops of the high-framerate video. This approach significantly boosts both processing speed and the resulting boundary detail. Finally, these high-detail mask segments are re-projected onto the full-frame canvas at their original coordinates, and the backwards process is run again to refine the initial frames of the new, high-resolution sequence.

\begin{figure}[!t]
    \centering
\includegraphics[width=\textwidth]{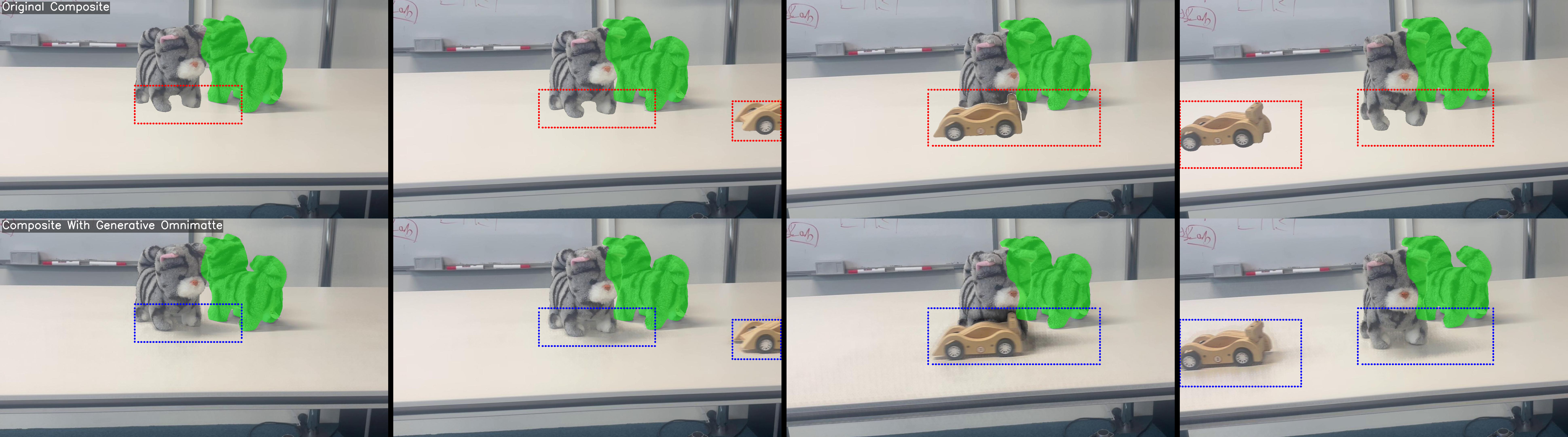}
    \caption{Comparison of data compositing with and without Generative Omnimatte. The bottom row, using our method, successfully captures secondary effects such as shadows and reflections. This results in a more realistic composite compared to the baseline method shown in the top row.}
    \label{fig:gen_omnimatte_effect}
\vspace{-0.7cm}
\end{figure}

\noindent\textbf{Compositing.}
With our high-resolution sequences, we use a multi-stage compositing pipeline to create the final dataset by realistically overlaying the foreground object, $I_\text{fg}$, onto the background scene, $I_\text{bg}$. We do so using a standard alpha blend with a soft alpha matte, $\alpha_\text{fg}\in [0,1]$, to allow for seamless, semi-transparent boundaries. The composite image, $I_\text{comp}$, is given by:
$
    I_\text{comp} = (1-\alpha_\text{fg})\cdot I_\text{bg} + \alpha_\text{fg}\cdot I_\text{fg}
$.
In practice, the alpha mattes are provided by Generative Omnimatte~\cite{lee2025generative}. This model captures both the foreground object and its secondary effects on the scene, such as cast shadows or reflections, and represents them in the soft alpha matte. Including these effects significantly increases the realism of the final composite, as shown in \cref{fig:gen_omnimatte_effect}. It is important to note that the soft alpha matte is used strictly for seamless compositing of boundaries. The interior of the occluder remains fully opaque ($\alpha_\text{fg}=1$), ensuring the background object is completely occluded in the final video data.

\subsection{Real-VAS Containment}\label{sec:Real-VAS Containment}

Our Container category is composed of three physically distinct scenarios (see \cref{fig:real-vas-fig}). While the initial interactions are unique, all three are built upon a unified principle for generating precise ground truth: once an object is fully contained, its motion is dictated by the motion of its container. To enable this, we enforce a physical constraint by using form-fitting object-container pairs. Computationally, this process relies on two key models. First, SAM2, is used to track an object while it is visible, establishing its last known mask before occlusion. Second, once the object is hidden, a point tracker, CoTracker3, captures the container's precise transformation. This transformation is then directly applied to the object's last known mask, treating them as a single rigid unit. A simpler alternative would be to propagate the object's mask by the container's frame-to-frame centroid displacement. This is insufficient in our setting: a centroid recovers only a translation vector, whereas a snugly-contained object inherits the container's full rigid or affine motion, including scale changes as the container moves toward or away from the camera, and rotation. CoTracker3 tracks multiple surface points on the container, from which we can estimate the full transform at sub-pixel precision.

Formally, let $O_t$ and $C_t$ denote the target-object and container masks at frame $t$, and let $t_\text{entry}$ be the frame at which the object becomes fully contained. During the visible phase ($t < t_\text{entry}$), $O_t$ is tracked directly with CoTracker3. At $t_\text{entry}$ the object's rigid boundary becomes tightly bounded by the inner topology of $C$, and the motion model transitions to
\begin{equation*}
    \mathbf{T}_{O_t} = \mathbf{T}_{C_t} + \epsilon,
    \qquad \forall\, t \geq t_\text{entry},
\end{equation*}
where $\mathbf{T}$ denotes the estimated rigid/affine transform and $\epsilon$ is a bounded local-deformation term whose magnitude we validate empirically in \cref{fig:container_val} of the supplementary (mean IoU $97.5\%$). When an external actuator (\eg a human hand) manipulates $O$ during insertion, we substitute the actuator's tracked translation for $\mathbf{T}_{C_t}$ until the object settles, at which point tracking transfers to $C$ (see \cref{fig:actuator_points_vis} in the supplementary).
The difference between scenarios lies in how these models are applied during the initial interaction:

\noindent\textbf{On Surface Containment:} inspired by the shell game, the object is static. The CoTracker3 logic is applied to the container from the moment of full occlusion.

\noindent\textbf{Articulated Containment:} we first use SAM2 to track the object's visible movement as it enters a fixture (\eg a drawer). Once occluded, we switch to using CoTracker3 on the container.

\noindent\textbf{Mobile Containment:} requires a multi-stage entry where we first reconstruct the partially occluded mask. To simulate the object settling, we infer the object's movement by tracking the visible actuator (\eg a human hand or tool) using CoTracker3 during the insertion phase. By assuming a fixed spatial relationship between the actuator and the object while it is being placed, we apply the tracked motion of the actuator directly to the object's last-known mask to determine its trajectory, speed, and final resting position within the container (see Supplementary Material). Only after the actuator releases the object and is removed do we engage the CoTracker3 logic on the container itself.

\subsection{Emulated Camera Motion}
\begin{figure}[!t]
    \centering
\includegraphics[width=\textwidth]{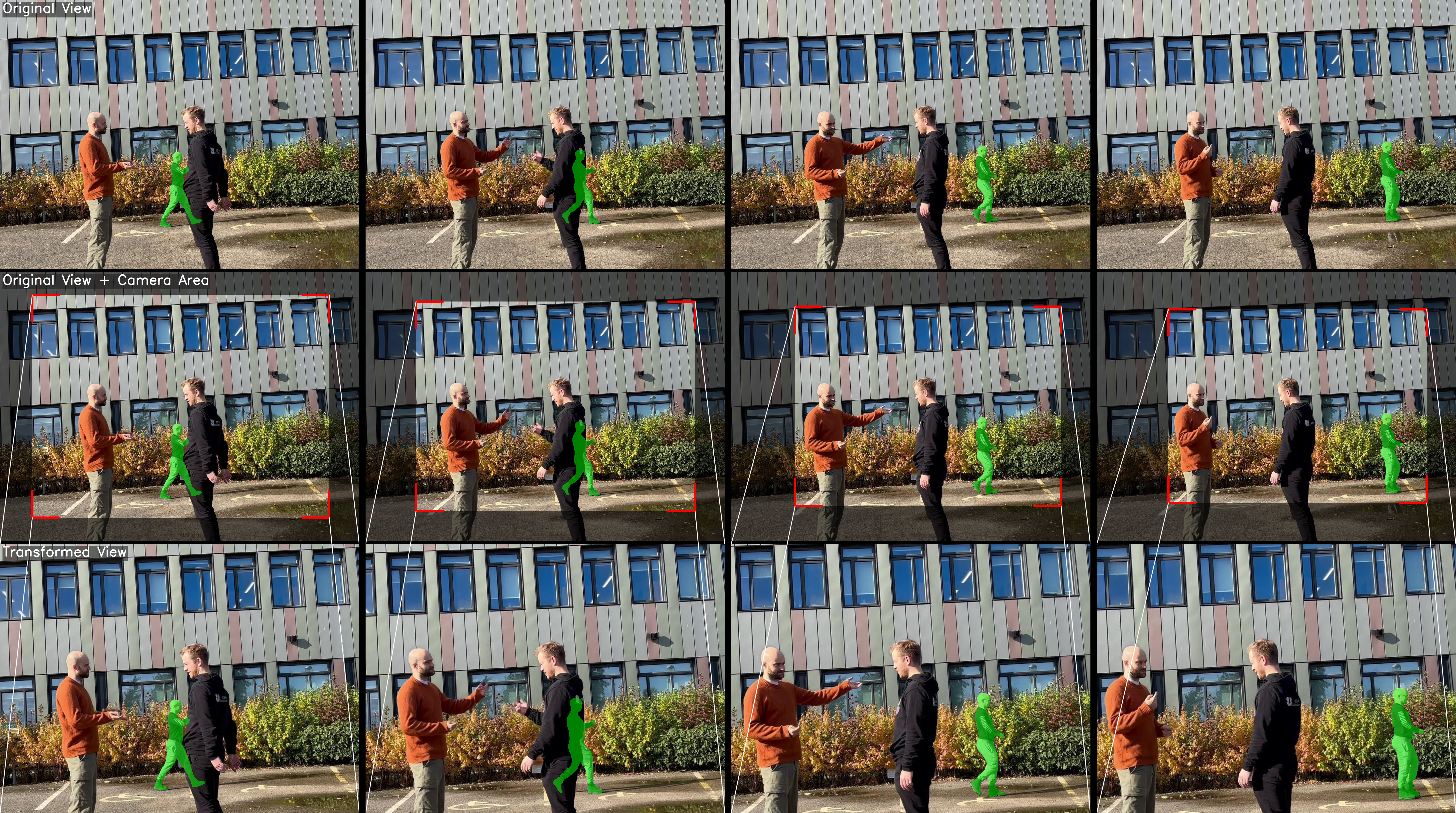}
    \caption{An example of emulated camera motion. This figure demonstrates a dynamic zoom and pan by showing the original frame, the specific sub-region that constitutes the new camera view, and the resulting transformed frame.}
    \label{fig:camera_motion}
    \vspace{-0.7cm}
\end{figure}

To ensure our dataset is as realistic and varied as possible, we emulate camera motion for scenes where the original data was collected with a static camera (\cref{fig:camera_motion}). This is mainly applied for our \textbf{Real-VAS Occlusion} set, which requires static camera data as input, whereas our container logic allows for camera motion. Our approach applies a diverse set of predefined camera motion profiles (\eg \textit{pans, zooms, and translations}) to the static source videos.

To ensure that the emulated motion does not corrupt the data by pushing key objects out of frame, we employ an \textbf{automated validation and selection pipeline}. Each potential motion profile is used to generate a candidate sequence, which is then evaluated against two critical criteria. First, we ensure the \textit{preservation of visibility}, verifying that the emulated motion does not push the primary objects (particularly the occludee) out of the frame. Second, we check for the \textit{preservation of occlusion quality}, ensuring that the interaction is maintained and the degree of occlusion is not substantially reduced from the original static version.
A suitable motion profile is then randomly selected from the pool of candidates that successfully pass these checks, introducing valuable diversity into our dataset. If no emulated motion satisfies our quality constraints for a given scene, the camera remains static, thereby preserving integrity.

\subsection{Dataset Statistics}

\cref{tab:dataset_comparison} presents a comparison between Real-VAS and other notable amodal segmentation datasets. Our contribution is significant in several key areas. First, Real-VAS provides precise annotations, removing the imprecision of human estimation found in other real-world datasets. Second, in terms of scale, it contains over 2.5 times as many frames as the next largest real-world benchmark. This scale is matched by its diversity, offering a substantially larger number of unique object classes. Finally, it is specifically designed for the amodal task, featuring both a higher occlusion prevalence and a much greater mean occlusion severity, making it a more challenging and suitable benchmark for evaluating amodal models. When compared to synthetic datasets, Real-VAS naturally trades sheer scale for real-world fidelity. While synthetic benchmarks like MOVi-MC-AC~\cite{moore2025training} offer a greater volume of scenes and classes, Real-VAS is intentionally curated to feature more challenging and realistic occlusions. This is evidenced by its mean occlusion severity, which is higher than that of most synthetic datasets. This focus on realism and the quality and difficulty of occlusions, rather than just the quantity, makes Real-VAS a suitable evaluation dataset for amodal segmentation. Further dataset metrics are presented in Appendix~\ref{sec:datasetscale}. We also present a scaling analysis (\cref{fig:metrics_per_ds_size}) in \cref{sec:scaling} that validates the sufficiency of this scale for stable benchmarking.

While our comparison in \cref{tab:dataset_comparison} is comprehensive, we have deliberately excluded some related datasets due to fundamental differences in their annotation formats. For instance, TAO-Amodal~\cite{hsieh2023tao}, despite being a large-scale real-world benchmark, is omitted as it only provides amodal annotations in the form of bounding boxes, which are not comparable to the pixel-level segmentation masks that our work focuses on. Similarly, the `Rubric' real-world evaluation set from TCOW~\cite{vanhoorick2023tcow} is excluded because it exclusively contains modal annotations (visible parts) of objects preceding and following an occlusion, and therefore lacks the necessary amodal ground truth to evaluate a model's performance during an occlusion event itself.

\subsection{Scaling Analysis}\label{sec:scaling}

\begin{figure}[!t]
    \centering
\includegraphics[width=\textwidth]{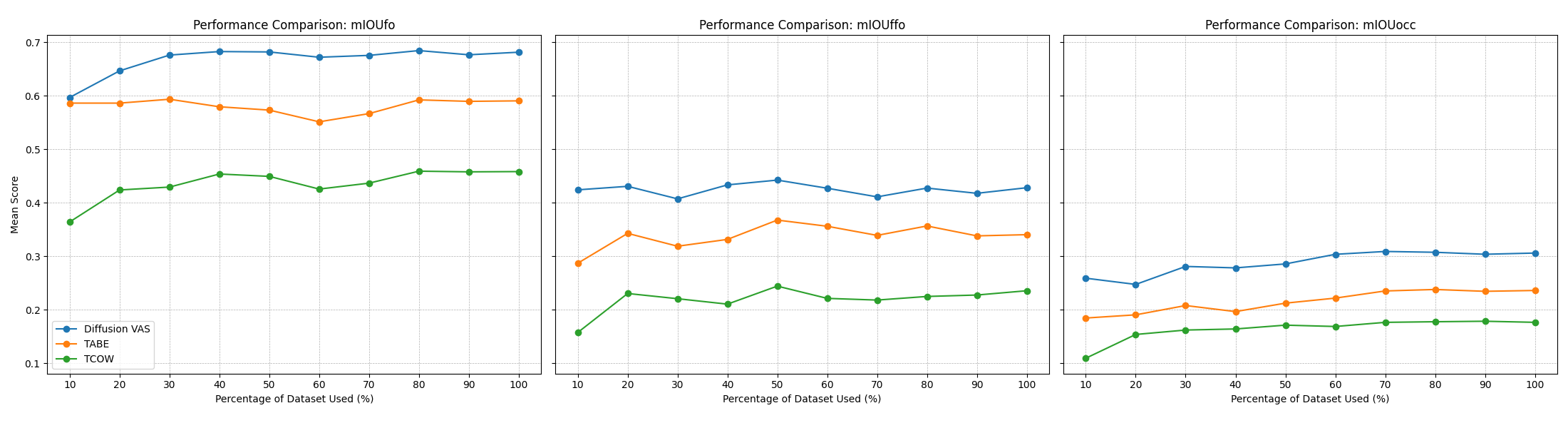}
    \caption{Scaling analysis of model performance on the Real-VAS dataset. The plots show the mean score for three different mIoU metrics as the percentage of the evaluation data is increased from 10\% to 100\%. The general convergence of the curves validates the dataset's scale as a stable and comprehensive benchmark.}
    \label{fig:metrics_per_ds_size}
\end{figure}

To validate Real-VAS as a robust evaluation benchmark, we performed a scaling analysis. We measured the performance of the top performing baseline methods on progressively larger random subsets of our test data, from 10\% to 100\%. The results (\cref{fig:metrics_per_ds_size}) demonstrate two key properties. First, the changing scores in the initial subsets (\eg from 10\% to 70\%) highlight the dataset's \textit{diversity}, showing that different parts of the dataset contain unique challenges that test various model capabilities. More importantly, as the evaluation size increases towards its full scale (from 80\% to 100\%), the performance scores for all models begin to stabilise and the curves flatten. This convergence is critical: it indicates that while an evaluation on a small subset might be sensitive to the specific scenes chosen, an evaluation on the full dataset provides a stable and reliable measure of a model's true performance. The fact that the scores stabilise demonstrates that Real-VAS is \textit{sufficiently large and comprehensive} to serve as a conclusive benchmark.

\section{Evaluation}
\label{sec:evaluation}

To benchmark performance on Real-VAS, we evaluated a diverse set of state-of-the-art (SOTA) models, as shown in \cref{tab:results}. These benchmarks include image-based models (pix2gestalt \cite{ozguroglu2024pix2gestalt}, SDAmodal \cite{zhan2024amodal}, PCNet \cite{zhan2020self}), video diffusion approaches (TABE \cite{hudson2024track}, Diffusion VAS \cite{chen2025using}, Generative Omnimatte \cite{lee2025generative}), and video transformers (TCOW \cite{vanhoorick2023tcow}). We excluded methods that output 3D amodal shapes from our benchmarks, such as Amodal3R~\cite{wu2025amodal3r}, as the lack of ground-truth camera intrinsics in our dataset prevents a fair evaluation of their projected 2D segmentation masks. Several of these frameworks require additional inputs beyond the initial query mask. For consistency, where a \textit{Modal Seg.} input is required, we provide visible-part segmentation masks generated by SAM2. \textit{BBox} indicates that the method requires a ground-truth bounding box to define the area for amodal completion. Despite this significant advantage, neither SDAmodal nor PCNet surpasses many methods lacking this extra information. We also distinguish between models by their \textit{Run Type}. Online methods process video sequentially and are suitable for real-time application, whereas offline methods process the entire clip at once, granting them far more context at the cost of higher latency. A significant performance trade-off exists between modelling paradigms. Video Transformer methods like TCOW offer substantially lower latency, often by orders of magnitude, compared to iterative Video Diffusion approaches such as TABE and Diffusion VAS.

\begin{table}[!t]
\centering
\resizebox{\textwidth}{!}{
\begin{tabular}{@{}lccccc@{}}
\toprule
\textbf{Method} & \textbf{Inputs} & \textbf{Run Type} & \textbf{$\text{mIoU}_\text{fo}$} & \textbf{$\text{mIoU}_\text{ffo}$} & \textbf{$\text{mIoU}_\text{occ}$} \\
\midrule
pix2gestalt \cite{ozguroglu2024pix2gestalt} & Modal Seg. & Online & 0.411 \scriptsize{$\pm 0.2167$} & 0.060 \scriptsize{$\pm 0.1074$} & 0.136 \scriptsize{$\pm 0.1088$} \\
TCOW \cite{vanhoorick2023tcow} & / & Offline & 0.426 \scriptsize{$\pm 0.2304$} & 0.179 \scriptsize{$\pm 0.1907$} & 0.165 \scriptsize{$\pm 0.1324$} \\
SDAmodal \cite{zhan2024amodal} & Modal Seg. \& GT BBox & Online & 0.526 \scriptsize{$\pm 0.2433$} & 0.059 \scriptsize{$\pm 0.1176$} & 0.146 \scriptsize{$\pm 0.1293$} \\
Generative Omnimatte \cite{lee2025generative} & Modal Seg. & Offline & 0.416 \scriptsize{$\pm 0.2194$} & 0.045 \scriptsize{$\pm 0.1284$} & 0.062 \scriptsize{$\pm 0.1002$} \\
TABE \cite{hudson2024track} & Modal Seg. & Offline & 0.501 \scriptsize{$\pm 0.2768$} & 0.209 \scriptsize{$\pm 0.2572$} & 0.207 \scriptsize{$\pm 0.1760$} \\
PCNet \cite{zhan2020self} & Modal Seg. \& GT BBox & Online & 0.452 \scriptsize{$\pm 0.2421$}  & 0.030 \scriptsize{$\pm 0.1114$} & 0.011 \scriptsize{$\pm 0.0435$} \\
Diffusion VAS \cite{chen2025using} & Modal Seg. & Offline & \textbf{0.604 \scriptsize{$\pm 0.2580$}} & \textbf{0.267 \scriptsize{$\pm 0.2944$}} & \textbf{0.262 \scriptsize{$\pm 0.2087$}} \\
\bottomrule
\end{tabular}
}
\caption{Quantitative results on complete Real-VAS dataset.}
\label{tab:results}
\vspace{-.5cm}
\end{table}

\begin{table}[t!]
\centering
\resizebox{\textwidth}{!}{%
\begin{tabular}{@{}lccccccc@{}}
\toprule
& \multicolumn{3}{c}{\textbf{Real-VAS Containment}} & & \multicolumn{3}{c}{\textbf{Real-VAS Occlusion}} \\
\cmidrule(r){2-4} \cmidrule(l){6-8}
\textbf{Method} & $\text{mIoU}_\text{fo}$ & $\text{mIoU}_\text{ffo}$ & $\text{mIoU}_\text{occ}$ & \phantom{a} & $\text{mIoU}_\text{fo}$ & $\text{mIoU}_\text{ffo}$ & $\text{mIoU}_\text{occ}$ \\
\midrule
pix2gestalt \cite{ozguroglu2024pix2gestalt} & 0.203 \scriptsize{$\pm 0.11$} & 0.016 \scriptsize{$\pm 0.03$} & 0.053 \scriptsize{$\pm 0.04$} & & 0.478 \scriptsize{$\pm 0.20$} & 0.096 \scriptsize{$\pm 0.13$} & 0.163 \scriptsize{$\pm 0.11$} \\
TCOW \cite{vanhoorick2023tcow} & 0.325 \scriptsize{$\pm 0.14$} & \textbf{0.108 \scriptsize{$\pm 0.14$}} & \textbf{0.128 \scriptsize{$\pm 0.11$}} & & 0.458 \scriptsize{$\pm 0.24$} & 0.236 \scriptsize{$\pm 0.21$} & 0.177 \scriptsize{$\pm 0.14$} \\
SDAmodal \cite{zhan2024amodal} & 0.335 \scriptsize{$\pm 0.19$} & 0.022 \scriptsize{$\pm 0.04$} & 0.076 \scriptsize{$\pm 0.05$} & & 0.587 \scriptsize{$\pm 0.23$} & 0.088 \scriptsize{$\pm 0.15$} & 0.168 \scriptsize{$\pm 0.14$} \\
Generative Omnimatte \cite{lee2025generative} & 0.250 \scriptsize{$\pm 0.15$} & 0.000 \scriptsize{$\pm 0.00$} & 0.003 \scriptsize{$\pm 0.01$} & & 0.470 \scriptsize{$\pm 0.21$} & 0.082 \scriptsize{$\pm 0.16$} & 0.081 \scriptsize{$\pm 0.11$} \\
TABE \cite{hudson2024track} & 0.225 \scriptsize{$\pm 0.14$} & 0.044 \scriptsize{$\pm 0.06$} & 0.118 \scriptsize{$\pm 0.09$} & & 0.590 \scriptsize{$\pm 0.25$} & 0.340 \scriptsize{$\pm 0.28$} & 0.236 \scriptsize{$\pm 0.19$} \\
PCNet \cite{zhan2020self} & 0.353 \scriptsize{$\pm 0.227$} & 0.013 \scriptsize{$\pm 0.02$} & 0.014 \scriptsize{$\pm 0.02$} &
& 0.533 \scriptsize{$\pm 0.19$} 
& 0.070 \scriptsize{$\pm 0.19$}  
& 0.009 \scriptsize{$\pm 0.05$}   \\
Diffusion VAS \cite{chen2025using} & \textbf{0.368 \scriptsize{$\pm 0.20$}} & 0.066 \scriptsize{$\pm 0.14$} & 0.125 \scriptsize{$\pm 0.13$} & & \textbf{0.681 \scriptsize{$\pm 0.23$}} & \textbf{0.428 \scriptsize{$\pm 0.29$}} & \textbf{0.306 \scriptsize{$\pm 0.21$}} \\
\bottomrule
\end{tabular}
}
\caption{Quantitative evaluation on Real-VAS, separated by Containment and Occlusion scenarios.}
\label{tab:results_per_split}
\vspace{-1.0cm}
\end{table}

Quantitative results are shown in \cref{tab:results}. A clear trend emerges where video-based methods (\eg TABE \cite{hudson2024track}, Diffusion VAS \cite{chen2025using}, TCOW \cite{vanhoorick2023tcow}) generally outperform image-based approaches (pix2gestalt \cite{ozguroglu2024pix2gestalt}, SDAmodal \cite{zhan2024amodal}, PCNet \cite{zhan2020self}). This is expected, as video models can leverage temporal context, which is particularly useful for interpolating across a full occlusion. A notable exception is the lower performance of Generative Omnimatte \cite{lee2025generative}; this result should be contextualised, as our evaluation protocol intentionally withholds the foreground (occluder) masks that the model is designed to use, thereby highlighting its dependency on that specific input. To further analyse model performance, \cref{tab:results_per_split} presents results split by our Occlusion and Containment scenarios. We observe a consistent drop in performance for all methods on the Containment tasks when compared to the simpler Occlusion events. This underscores the difficulty of these interactive scenarios and confirms the importance of including them in a comprehensive amodal segmentation benchmark. It is noteworthy, however, that TCOW outperforms all other methods on Containment $\text{mIoU}_\text{ffo}$ and $\text{mIoU}_\text{occ}$, this is likely attributable to its training on synthetic containment-style videos, which demonstrates that performance on these complex physical reasoning tasks is heavily influenced by exposure to similar training data.
To further investigate cross-dataset generalisation and the sim2real gap, we evaluated the transferability of models trained on synthetic data to our real-world benchmark. Using Diffusion VAS as a case study, we observed a substantial performance degradation. When evaluated on the large-scale synthetic SAIL-VOS dataset \cite{hu2019sail}, Diffusion VAS achieved a reported $mIoU_{occ}$ of 0.55 \citep{chen2025using}. However, when this same model is evaluated on the Real-VAS Occlusion split, performance drops significantly to 0.31 $mIoU_{occ}$. The degradation is even more severe on the Real-VAS Containment split, where the model achieves just 0.13 $mIoU_{occ}$. This substantial performance gap highlights that Real-VAS captures distinct, real-world challenges and complex physical interactions that are fundamentally absent from existing synthetic benchmarks, confirming the necessity of a real-world evaluation dataset.

Beyond evaluation, Real-VAS also provides useful training signal: fine-tuning Diffusion VAS on a random 300-sequence subset improves both in-domain performance and cross-domain generalisation to TAO-Amodal (see \cref{sec:train_util}).
\section{Discussion}
\label{sec:conclusion}

In this work, we addressed the critical challenge of creating a large-scale, high-fidelity dataset for video amodal segmentation. We introduced Real-VAS, a novel dataset generated through a sophisticated pipeline that leverages real-world footage to produce programmatically consistent ground-truth annotations, avoiding the independent annotation error found in manually-labelled real-world benchmarks. This approach successfully bridges the sim2real gap found in purely synthetic data while bypassing the inaccuracies and high costs associated with manual annotation or 3D reconstruction. A key contribution of our work is the introduction of the Container scenarios, which push beyond the simple pass-by occlusions prevalent in existing datasets. By providing structured, physically-grounded interactions such as On Surface Containment, Articulated Containment, and Mobile Containment, Real-VAS establishes a new and challenging benchmark for evaluating a model's capacity for physical reasoning and object permanence. The inclusion of both Occlusion and Container scenarios is deliberate, as we hypothesise they require different model capabilities. For instance, the structured logic of Containment may be solvable by models that excel at physical or relational reasoning, yet these same models might fail at the visual completion needed for unstructured Occlusion. Conversely, powerful generative models that can hallucinate appearance for Occlusion events may lack the explicit physical constraints required to succeed at Containment. Our dataset is designed to spur the development of unified solutions that can solve both. Our pipeline further enhances realism by emulating camera motion and incorporating secondary effects like cast shadows.

\paragraph{Scope and Limitations.}

Real-VAS is a \emph{composited real-image} benchmark: the source footage is real, and camera noise, texture and sensor artefacts are preserved, but occlusion events are synthesised by compositing rather than physically recorded. This design choice lets us scale programmatically consistent amodal ground truth without 3D-capture rigs or subjective human annotation of hidden regions, at the cost of occasional lighting inconsistencies or contact-point artefacts. Our Container scenarios further rely on a bounded ``snug-fit'' proxy constraint, which does not cover loose containment, such as a small object moving freely inside a large box. Future work could extend the generation pipeline to include controlled real-world camera motion, fluid dynamics, and more complex self-occlusion.

\bibliographystyle{splncs04}
\bibliography{refs}

\appendix
\newpage
\section{Further Dataset Generation Details}
\label{sec:ds_gen_details}

\subsection{Segmentation Class List}\label{sec:classlist}
To identify objects that are inherently dynamic, we define a comprehensive set of `moving object' classes. This set is broadly categorized to cover a wide range of potential motion.

It includes biological entities such as person, animal, bird, dog, cat, fish, horse, and insect. It also features a wide array of vehicles, from terrestrial (car, bus, motorcycle, train) and marine (boat, ship) to aerial (airplane, helicopter, drone). Furthermore, we include mechanical systems and devices known for their movement, like robot, conveyor belt, fan, and clock hands. Finally, the list is rounded out by common recreational and everyday objects whose behavior is often dynamic, including ball, kite, swing, and shopping cart.

\subsection{Use of CoTracker3 Within Containment Logic}
For our ground-truth generation in containment scenarios, we found that running CoTracker3 in online mode was significantly more robust than in offline mode. This is because the offline tracker, which processes the entire video at once, was prone to identity switching. In cases where multiple identical containers crossed paths, the offline model would incorrectly assign the track to a different container that later occupied the same space. The online, frame-by-frame approach mitigates this issue by maintaining a continuous track without being influenced by future events.

\subsection{Combining Monocular Depth with Segmentation}
\begin{figure}[!h]
    \centering
\includegraphics[width=\textwidth]{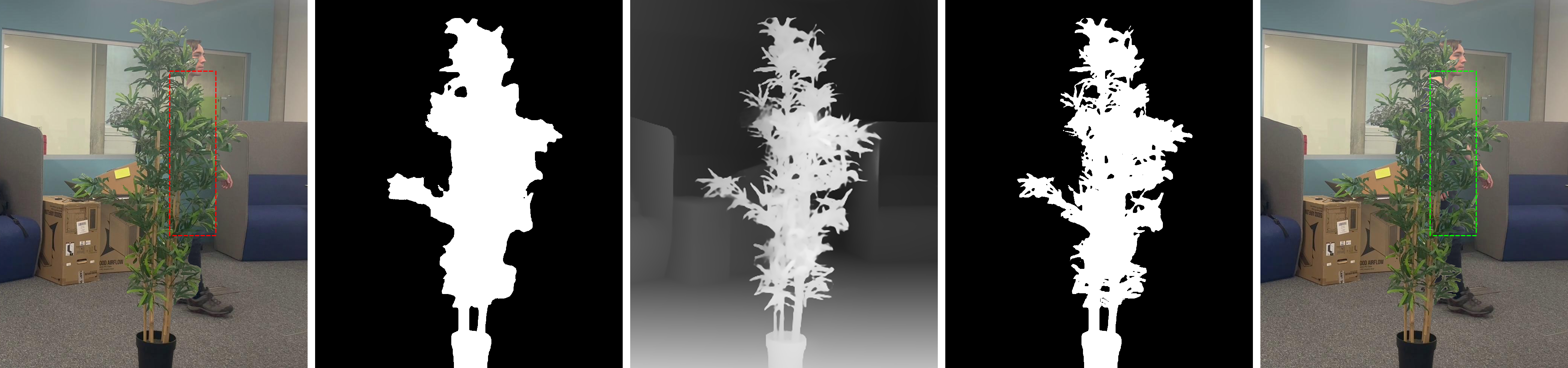}
    \caption{Refining a segmentation mask using monocular depth. From left to right: (a) The initial composite showing the person occluded by the plant (missing upper torso and upper leg). (b) The coarse mask from SAM2. (c) The corresponding depth map. (d) The refined mask after using depth to exclude background pixels. (e) The final, improved composite, where details of the person are correctly revealed through the gaps in the plant.}
    \label{fig:monodepth_seg_example}
\end{figure}

In cases where segmentation (SAM2) struggles with fine details, we leverage monocular depth (Depth Anything V2) to refine compositing. By calculating the average depth within the mask, we can exclude regions with greater depth values, effectively isolating the background. This technique substantially improves the realism of our composites. As shown in \cref{fig:monodepth_seg_example}, it allows for visibility through intricate structures, like the gaps between the leaves of a plant, correctly revealing the parts of the person.

\subsection{Dataset Generation Speed and Computation Requirements}
Our data generation pipeline was executed on an NVIDIA A40 GPU. While it is not strictly a requirement, a GPU with a large VRAM is beneficial for processing long video sequences, where the computational demands of the SAM2 mask propagation step can be significant.

Quantifying the precise generation speed is challenging due to the highly parallelised and automated nature of the pipeline. In particular, the clip alignment stage, where our system searches for plausible occlusion events, was designed to run overnight on large batches of videos to maximise the number of high-quality candidates discovered.

\subsection{Manual Quality Control Verification Stages}\label{sec:qc}
While our data generation pipeline is fully automated, we incorporate two quality control checks on the videos processed in the Real-VAS dataset to ensure the final dataset meets a high standard of quality and accuracy required for an evaluation dataset. We have built tools to ensure these QC checks are highly efficient.

\begin{figure}[!t]
    \centering
\includegraphics[width=\textwidth]{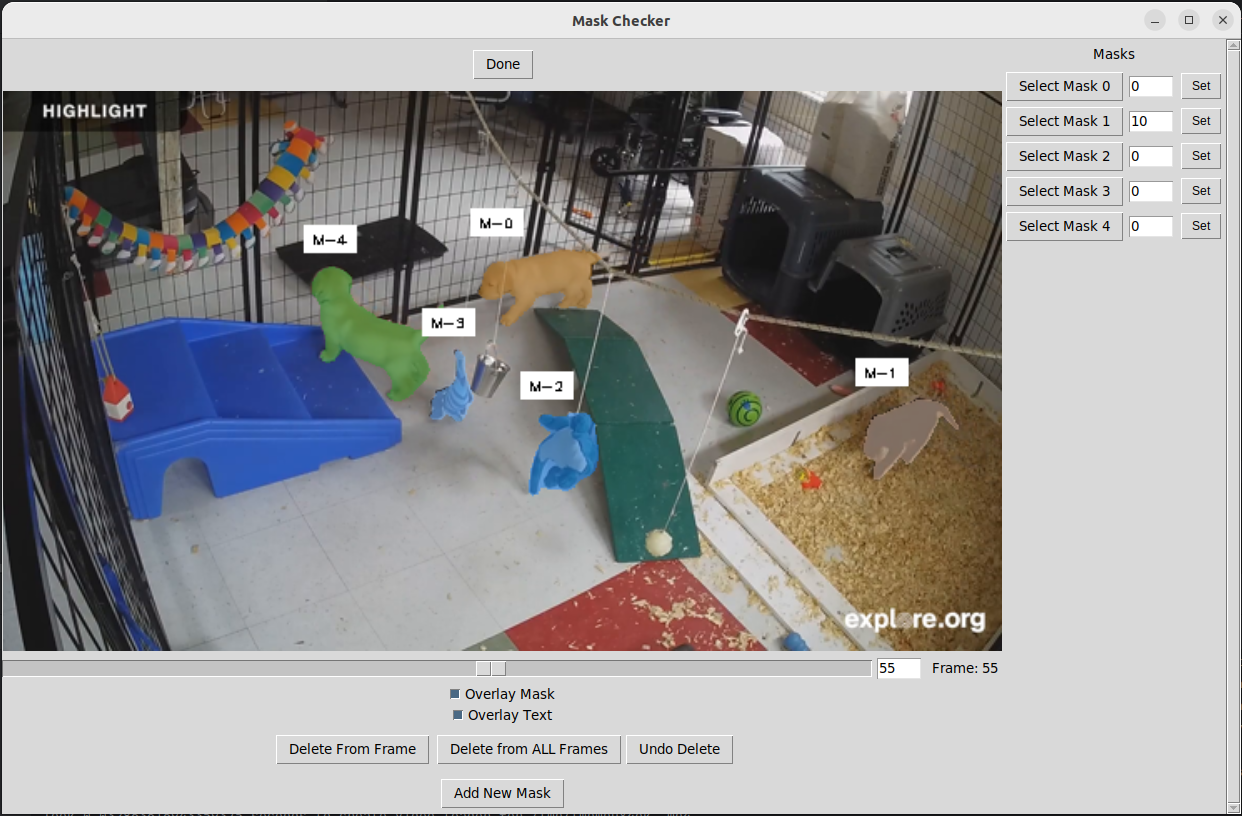}
    \caption{The GUI for the mask verification step. This tool allows users to easily correct inaccuracies by adding, modifying, or removing masks.}
    \label{fig:mask_checker_gui}
\end{figure}

First, following the automated mask generation, we perform a rapid curation step. Using a lightweight, purpose-built GUI (\cref{fig:mask_checker_gui}), human curators can quickly cycle through masks to correct obvious segmentation errors or add any that were missed. This interface is designed for speed, turning a potentially laborious task into a quick quality check that focuses only on refining clear inaccuracies.

\begin{figure}[!t]
    \centering
\includegraphics[width=\textwidth]{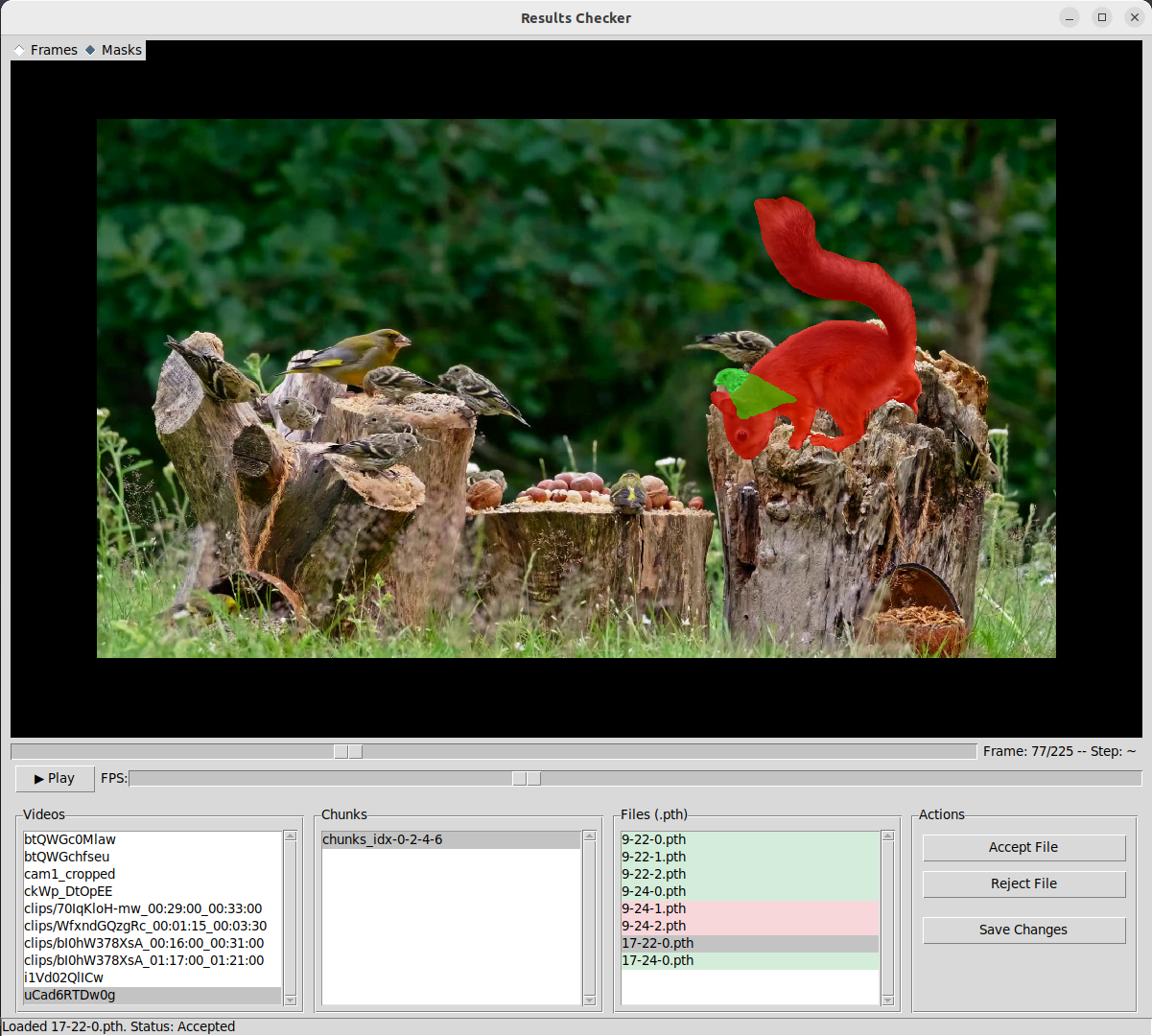}
    \caption{Our simple GUI for the final verification step, where generated clips are quickly accepted or rejected based on their overall quality.}
    \label{fig:final_manual_verifcation}
\end{figure}

Second, after a full sequence has been composited, it undergoes a final acceptance review. Curators use a simple viewing tool to watch the short clip and give a quick pass/fail judgement based on its overall realism and interest, as shown in \cref{fig:final_manual_verifcation}. This final check is extremely fast and ensures that only the most compelling and accurate sequences are included in the dataset. By designing these lightweight interfaces, our human-in-the-loop stages serve as an efficient quality control mechanism rather than a time-consuming annotation bottleneck.

To quantify the fidelity of the resulting masks, we additionally re-annotated a random sample of 500 frames drawn from 50 sequences in CVAT and compared the manual masks against those produced by our pipeline. The mean per-sequence IoU is $0.966$, with a minimum of $0.947$, confirming high fidelity on the source segmentation that underlies the composited ground truth.

\section{Ablation Studies}
\label{sec:ablations}
\subsection{Precision of Containment Logic}
A key challenge for our Containment scenarios is that once an object is hidden, its true position cannot be visually verified. To quantitatively validate the precision of our ground-truth generation logic, we therefore conducted a specialised experiment using videos of transparent containers.

This validation process involves two steps. First, we establish a verifiable ground truth by running a standard segmentation model on the original transparent videos. Since the object is always visible through the container, this provides its true mask for every frame. Second, we digitally render the transparent containers opaque in these same videos and then run our proposed containment logic. This logic takes the object's last visible mask before it becomes hidden and uses a point tracker on the (now opaque) container to predict the object's subsequent amodal masks.

\begin{figure}[!t]
    \centering
\includegraphics[width=\textwidth]{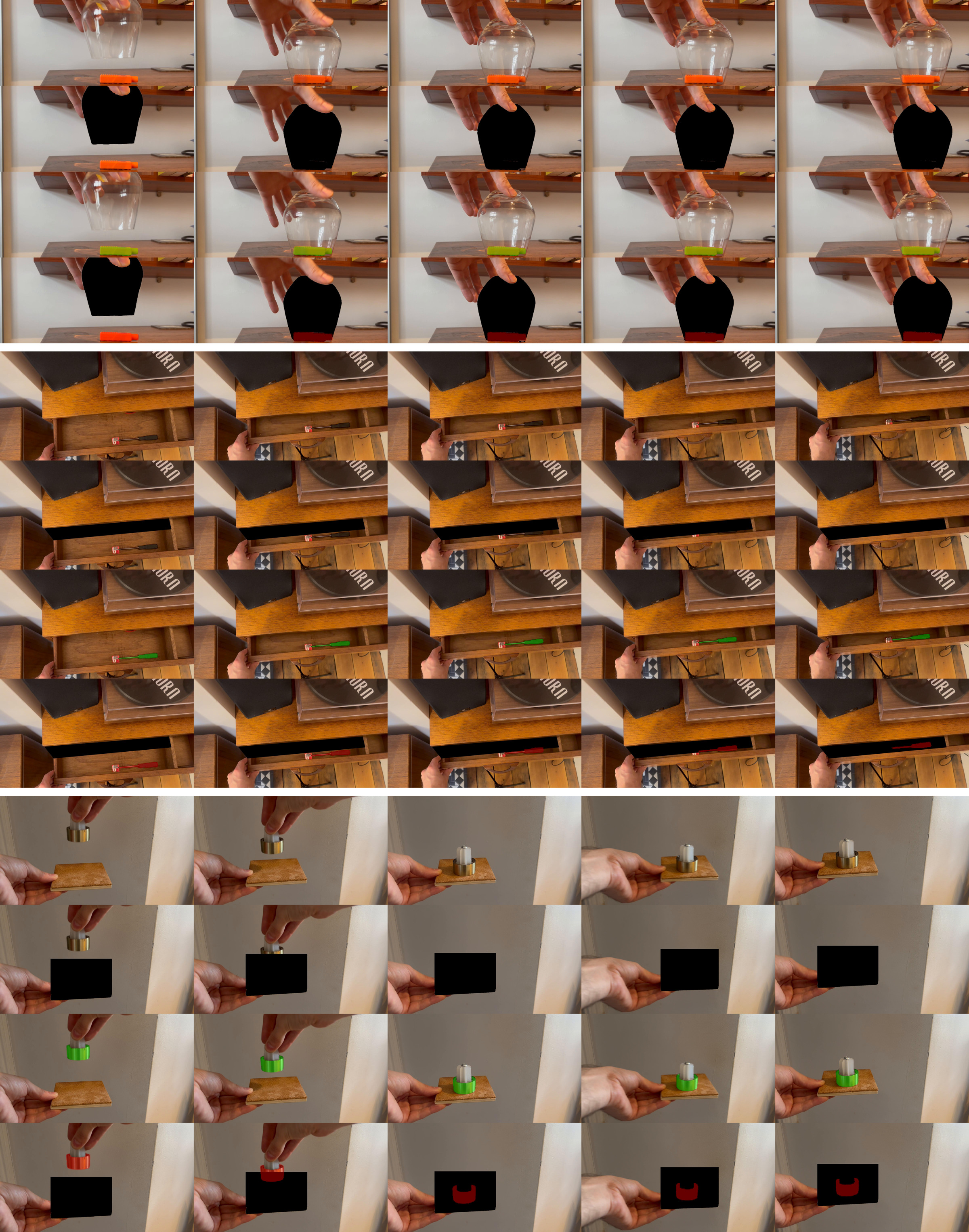}
    \caption{Validation of our containment logic across three key scenarios: On-Surface, Articulated, and Mobile Containment. For each example, the rows display (from top to bottom): the original frame, the synthetically applied container occlusion, the ground truth mask, and the final mask predicted by our logic. Our approach is highly accurate, achieving a 97.5\% Intersection over Union (IoU) with the ground truth masks.}
    \label{fig:container_val}
\end{figure}

Across all our container scenarios, comparing our logic's predicted masks against the verifiable ground truth yields a mean IoU of 97.5\%. While this score is very high, we note that IoU is highly sensitive to minor pixel shifts, and the segmentation masks used as a reference are not perfectly noise-free. Nevertheless, this result demonstrates that our containment logic is highly precise. We showcase the experimental setup for each scenario in \cref{fig:container_val}.

\subsection{Pipeline Component Ablation}\label{sec:pipeline_ablation}
To quantify how each stage of our generation pipeline contributes to benchmark difficulty, we ablate two components, emulated camera motion (\cref{fig:camera_motion}) and Generative Omnimatte compositing (\cref{fig:gen_omnimatte_effect}), and re-evaluate two representative models. Because the pipeline produces evaluation data rather than model outputs, we treat zero-shot performance as a proxy for realism: components that increase realism should make the task harder.

\cref{tab:pipeline_ablation} confirms this. Removing camera motion produces the largest single gain for TCOW, as the model no longer needs to disentangle ego-motion from true object displacement. Removing Generative Omnimatte accounts for a further, independent gain, because the resulting composites have harsher artificial boundaries and lack the cast shadows and reflections that would otherwise prevent models from exploiting simple texture discontinuities. Removing both components is easiest of all for both models, confirming that they contribute independently to difficulty.

\begin{table}[h!]
\centering
\small
\begin{tabular}{l cc}
\toprule
Variant & TCOW & Diffusion VAS \\
        & $\text{mIoU}_\text{fo}$ & $\text{mIoU}_\text{fo}$ \\
\midrule
Real-VAS (full)          & 0.426 & 0.604 \\
w/o Generative Omnimatte & 0.524 & 0.619 \\
w/o camera motion        & 0.547 & 0.613 \\
w/o both                 & 0.560 & 0.628 \\
\bottomrule
\end{tabular}
\caption{Pipeline component ablation. Both emulated camera motion and Generative Omnimatte compositing independently contribute to benchmark difficulty.}
\label{tab:pipeline_ablation}
\end{table}

\subsection{Training Utility}\label{sec:train_util}
While Real-VAS is designed primarily as an evaluation benchmark, we verify that it also provides useful training signal. We fine-tune Diffusion VAS~\cite{chen2025using} on a random 300-sequence subset of Real-VAS and evaluate on (i) the held-out 100-sequence test split, and (ii) TAO-Amodal~\cite{hsieh2023tao} zero-shot. Results are shown in \cref{tab:finetune}. Fine-tuning yields improvements on both in-domain and cross-domain evaluations. In particular the AP$_{75}$ gain on TAO-Amodal ($0.720 \rightarrow 0.742$), a strict metric requiring high-overlap masks, indicates that the precise composited masks in Real-VAS teach the model tighter output estimates than those learned from the much larger synthetic SAIL-VOS~\cite{hu2019sail} corpus used during pre-training. Achieving these gains from only 300 sequences suggests Real-VAS transfers substantially more efficiently than existing synthetic training data, and that scaling the pipeline offers a viable path to a large-scale real-world training set.

\begin{table}[h!]
\centering
\small
\begin{tabular}{l ccc | cc}
\toprule
 & \multicolumn{3}{c|}{TAO-Amodal (Top-1)} & \multicolumn{2}{c}{Real-VAS} \\
 & AP$_{25}$ & AP$_{50}$ & AP$_{75}$ & $\text{mIoU}_\text{fo}$ & $\text{mIoU}_\text{occ}$ \\
\midrule
Baseline   & 0.973 & 0.893 & 0.720 & 0.628 & 0.288 \\
Fine-tuned & \textbf{0.974} & \textbf{0.899} & \textbf{0.742} & \textbf{0.652} & \textbf{0.293} \\
\bottomrule
\end{tabular}
\caption{Fine-tuning Diffusion VAS on 300 randomly-selected Real-VAS sequences improves both in-domain performance (100-sequence held-out split) and cross-domain generalisation to TAO-Amodal.}
\label{tab:finetune}
\end{table}

\subsection{Performance vs.\ Occlusion Duration}\label{sec:duration}
\begin{figure}[!t]
    \centering
    \includegraphics[width=\textwidth]{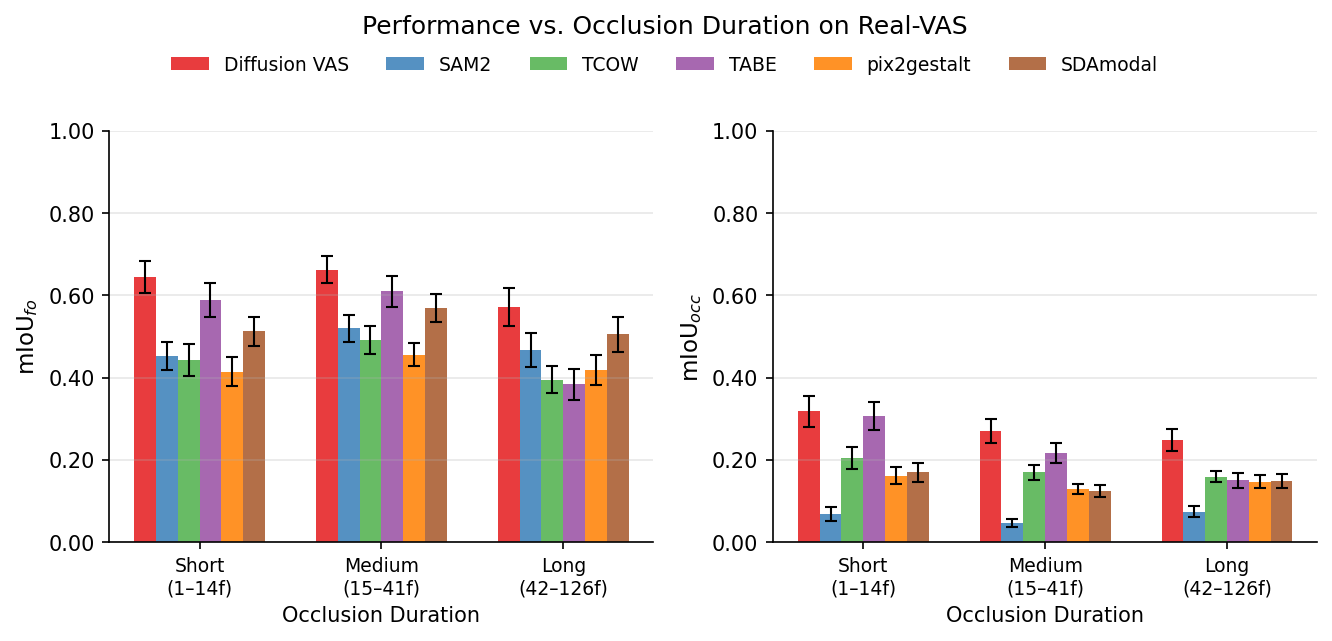}
    \caption{Performance vs.\ occlusion duration across SOTA models on Real-VAS. Clips are binned by duration into Short (1--14 frames), Medium (15--41), and Long (42--132). Every method's $\text{mIoU}_\text{fo}$ drops from Medium to Long occlusions, validating that Real-VAS tests object permanence over extended periods rather than a bias towards short pass-by events.}
    \label{fig:duration}
\end{figure}

To verify that Real-VAS stresses long-horizon object permanence rather than short pass-by occlusions, we bin every evaluation clip by occlusion duration into Short (1--14 frames), Medium (15--41), and Long (42--132) and re-evaluate all baselines. \Cref{fig:duration} shows the resulting $\text{mIoU}_\text{fo}$ scores within each bin. Every method's $\text{mIoU}_\text{fo}$ drops from Medium to Long occlusions, confirming that the difficulty of Real-VAS is driven by sustained occlusion rather than an imbalance in short pass-by events.

\subsection{Further Dataset Statistics}\label{sec:datasetscale}
In \cref{tab:further_ds_stats} we showcase some additional statistics of our Real-VAS dataset supplementing the comparisons in \cref{tab:dataset_comparison}. These metrics highlight the dataset's diversity, showing a wide variation in clip lengths, frame rates and resolutions. Furthermore, \cref{tab:clip_split} demonstrates good coverage across our Occlusion and Container scenarios.

\subsection{Use of CoTracker3 On Actuators Within Containment Logic}\label{sec:actuator_tracking}
\begin{figure}[!t]
    \centering
\includegraphics[width=\textwidth]{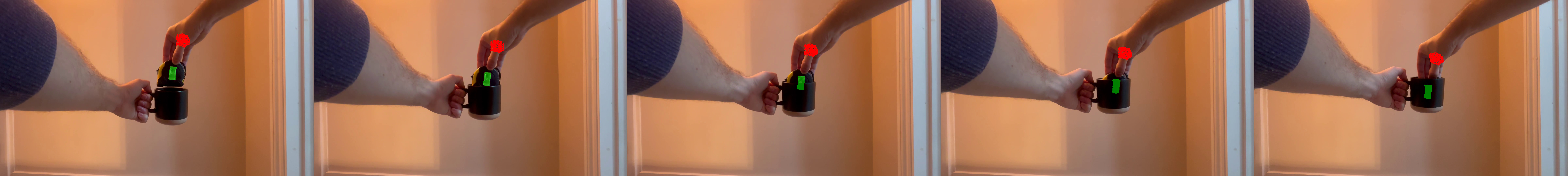}
    \caption{A visualisation of the actuator tracking mechanism used during the Mobile Containment phase. The red point indicates the tracked location on the actuator (the human hand), while the green mask represents the object being placed into the container. By tracking the hand's motion, we accurately simulate the object's settling trajectory as it becomes fully occluded.}
    \label{fig:actuator_points_vis}
\end{figure}

As discussed in \cref{sec:Real-VAS Containment}, determining the precise resting position of an object during Mobile Containment requires tracking the actuator that places it. \Cref{fig:actuator_points_vis} illustrates this multi-stage settling process. We initialise a point track (visualised as a red dot) on a consistently visible portion of the actuator, in this case, the human hand, using CoTracker3. By assuming a rigid spatial relationship between the actuator and the object during the insertion phase, the frame-to-frame displacement of this tracked point is applied directly to the object's last known segmentation mask (shown in green). This methodology ensures that the object's inferred trajectory, speed, and final settling position inside the container are physically grounded in visual evidence rather than relying on arbitrary physics simulations.

\begin{table}[h!]
\centering
\begin{tabular}{ccc | ccc | ccc}
\toprule
\multicolumn{3}{c}{\textbf{Clip Length (frames)}} & \multicolumn{3}{c}{\textbf{Frames Per Second}} &
\multicolumn{3}{c}{\textbf{Resolution}}\\
\cmidrule(lr){1-3} \cmidrule(lr){4-6} \cmidrule(lr){7-9}
\textbf{Min} & \textbf{Max} & \textbf{Avg} & \textbf{Min} & \textbf{Max} & \textbf{Avg} &
\textbf{Min} & \textbf{Max} & \textbf{Mode} \\
\midrule
3 & 132 & 37 & 10 & 60 & 16 & $720\times 1280$ & $2160\times 3840$ & $1080\times 1920$ \\
\bottomrule
\end{tabular}
\caption{Further statistics of our Real-VAS Dataset.}\label{tab:further_ds_stats}
\end{table}

\begin{table}[h!]
\centering
\begin{tabular}{c | cccc}
\toprule
\textbf{Num.\ Occlusion Clips} & \multicolumn{4}{c}{\textbf{Num.\ Containment Clips}} \\
\cmidrule(lr){2-5}
\ & \textbf{Total} & \textbf{On Surface} & \textbf{Articulated} & \textbf{Mobile} \\
\midrule
 300 & 100 & 38 & 33 & 29 \\
\bottomrule
\end{tabular}
\caption{The distribution of clips between our Occlusion and Container categories. The number of container scenarios is smaller by design, reflecting the inherent difficulty of sourcing the specific, high-quality physical interactions required for this category. In contrast, simple occlusion events are far more common in in-the-wild footage, allowing for a greater volume and variety.}\label{tab:clip_split}
\end{table}

\begin{figure}[tb]
    \centering
\includegraphics[width=\textwidth]{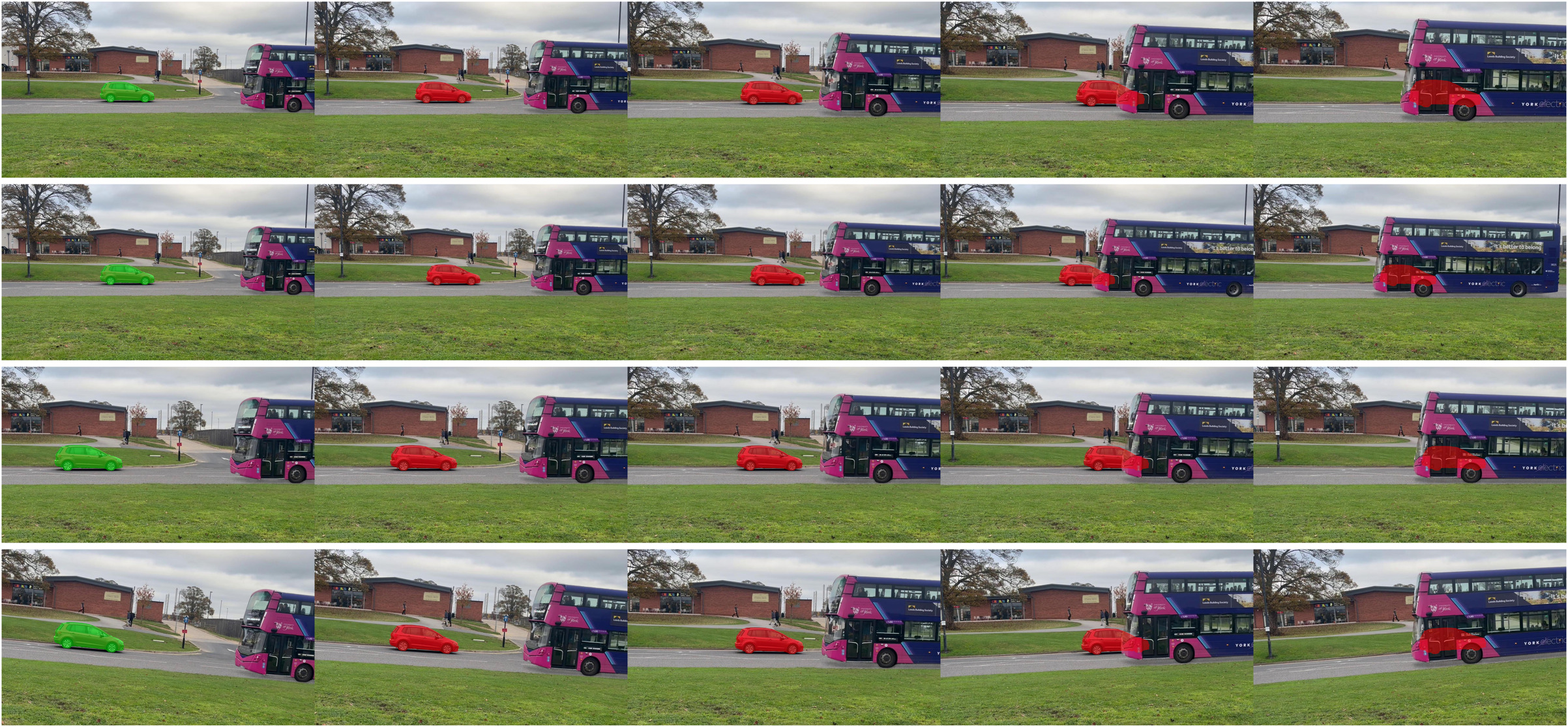}
    \caption{A selection of example outputs from our camera emulation module. The figure displays four profiles from our larger library of predefined motions, which are applied to static videos to enhance realism and introduce challenging, dynamic viewpoints to the dataset.}
    \label{fig:CameraMotionExample}
\end{figure}

\begin{figure}[tb]
    \centering
\includegraphics[width=\textwidth]{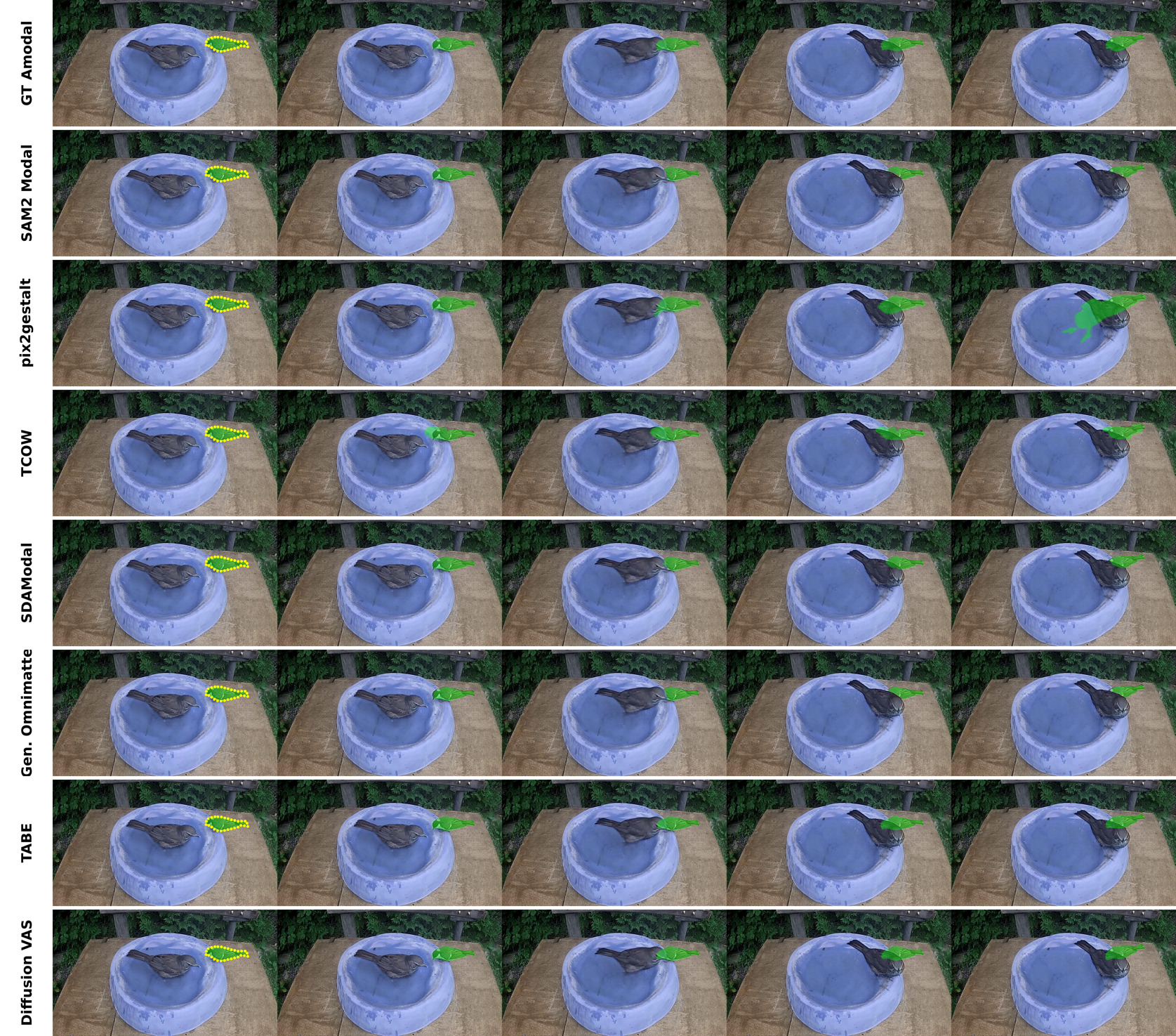}
    \caption{Qualitative outputs from each method -- Part 1.}
    \label{fig:QualExamples}
\end{figure}

\begin{figure}[tb]
    \centering
\includegraphics[width=\textwidth]{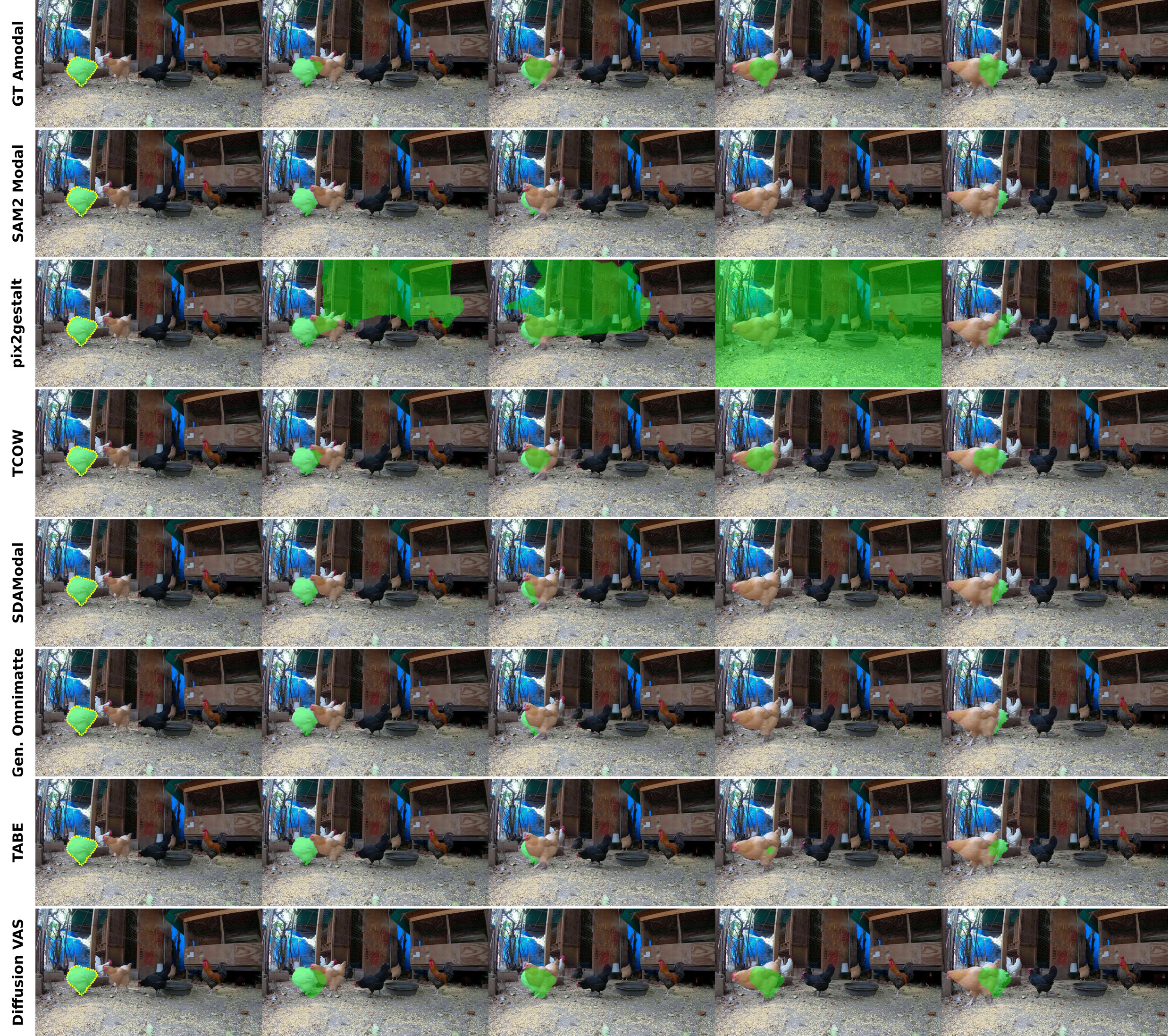}
    \caption{Qualitative outputs from each method -- Part 2.}
    \label{fig:QualExamples2}
\end{figure}

\begin{figure}[tb]
    \centering
\includegraphics[width=\textwidth]{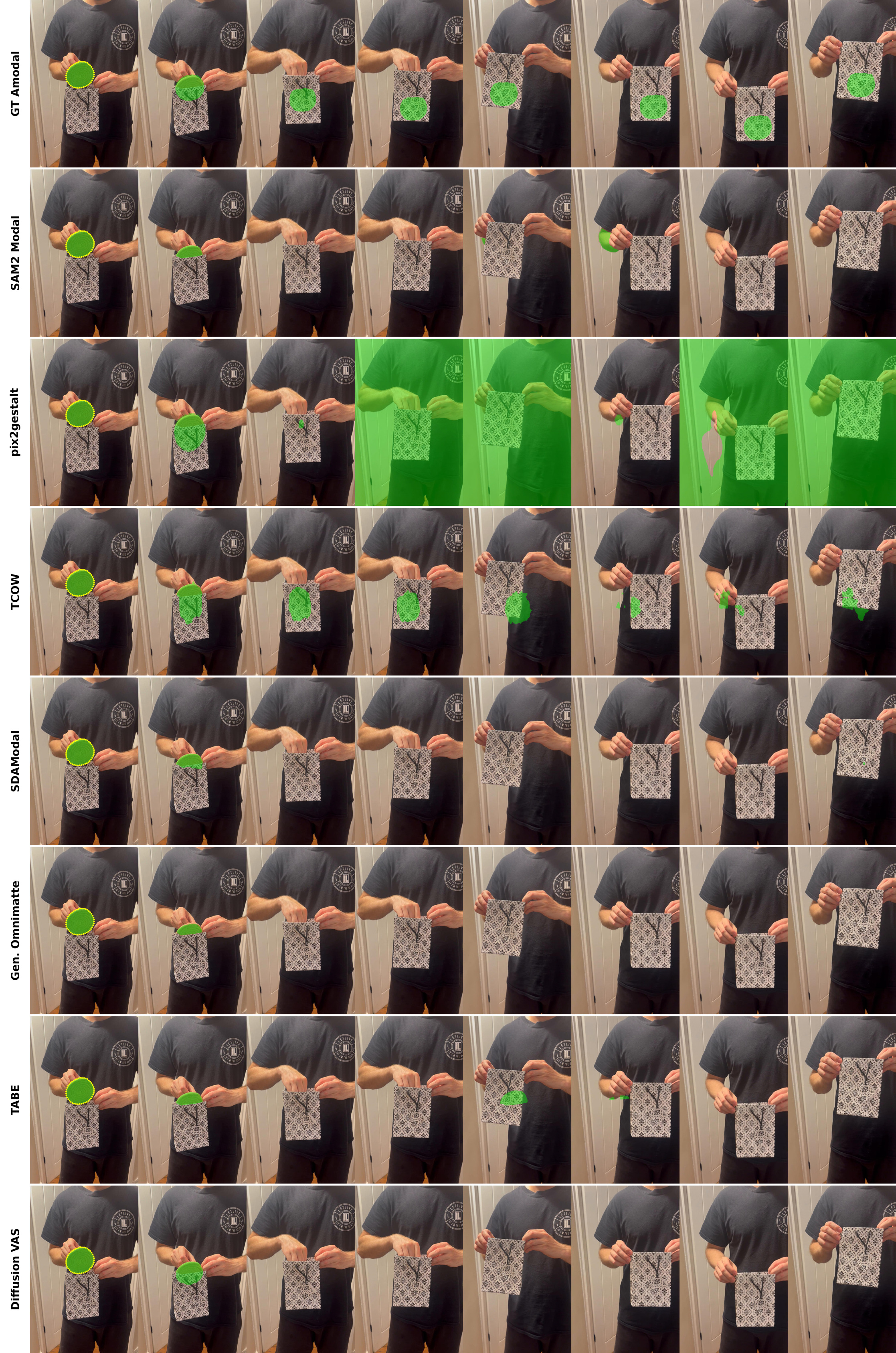}
    \caption{Qualitative outputs from each method -- Part 3.}
    \label{fig:QualExamples3}
\end{figure}

\begin{figure}[tb]
    \centering
\includegraphics[width=\textwidth]{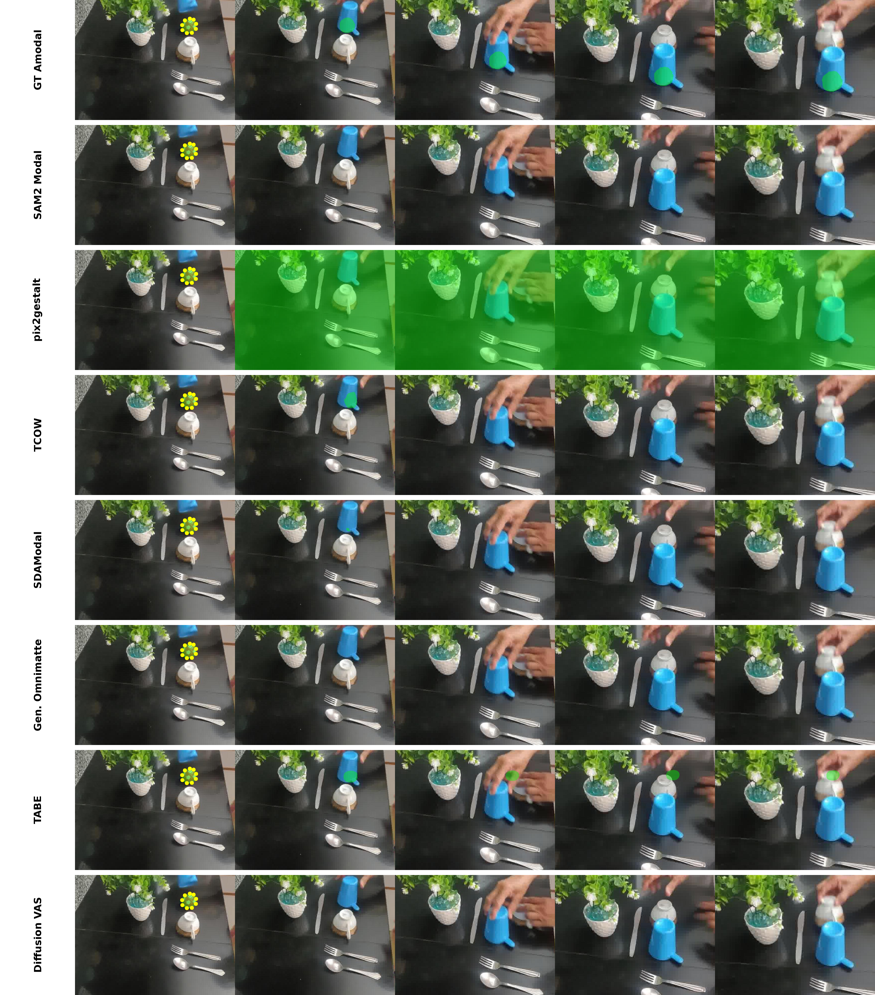}
    \caption{Qualitative outputs from each method -- Part 4.}
    \label{fig:QualExamples4}
\end{figure}

\begin{figure}[tb]
    \centering
\includegraphics[width=\textwidth]{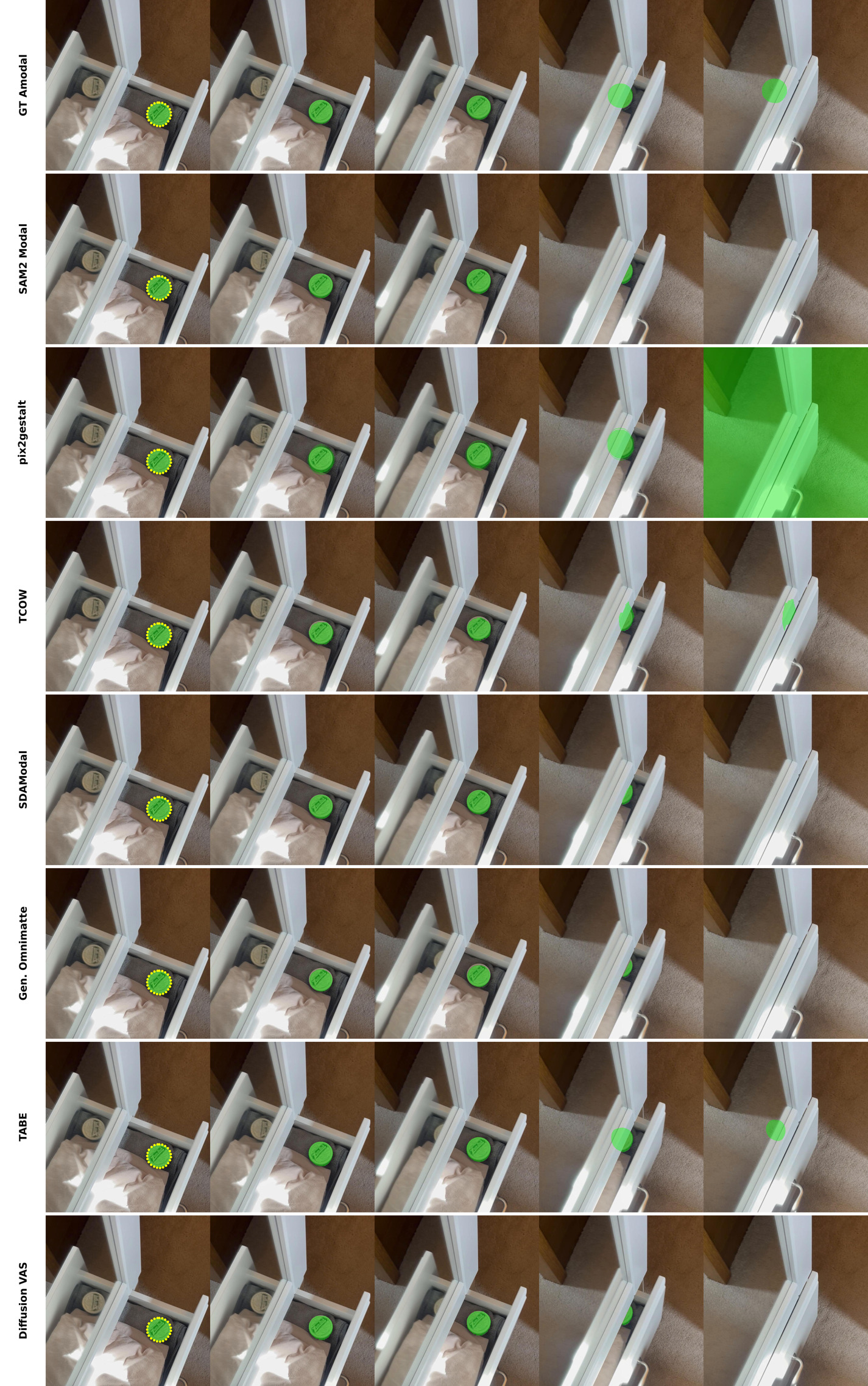}
    \caption{Qualitative outputs from each method -- Part 5.}
    \label{fig:QualExamples5}
\end{figure}

\begin{figure}[tb]
    \centering
\includegraphics[width=\textwidth]{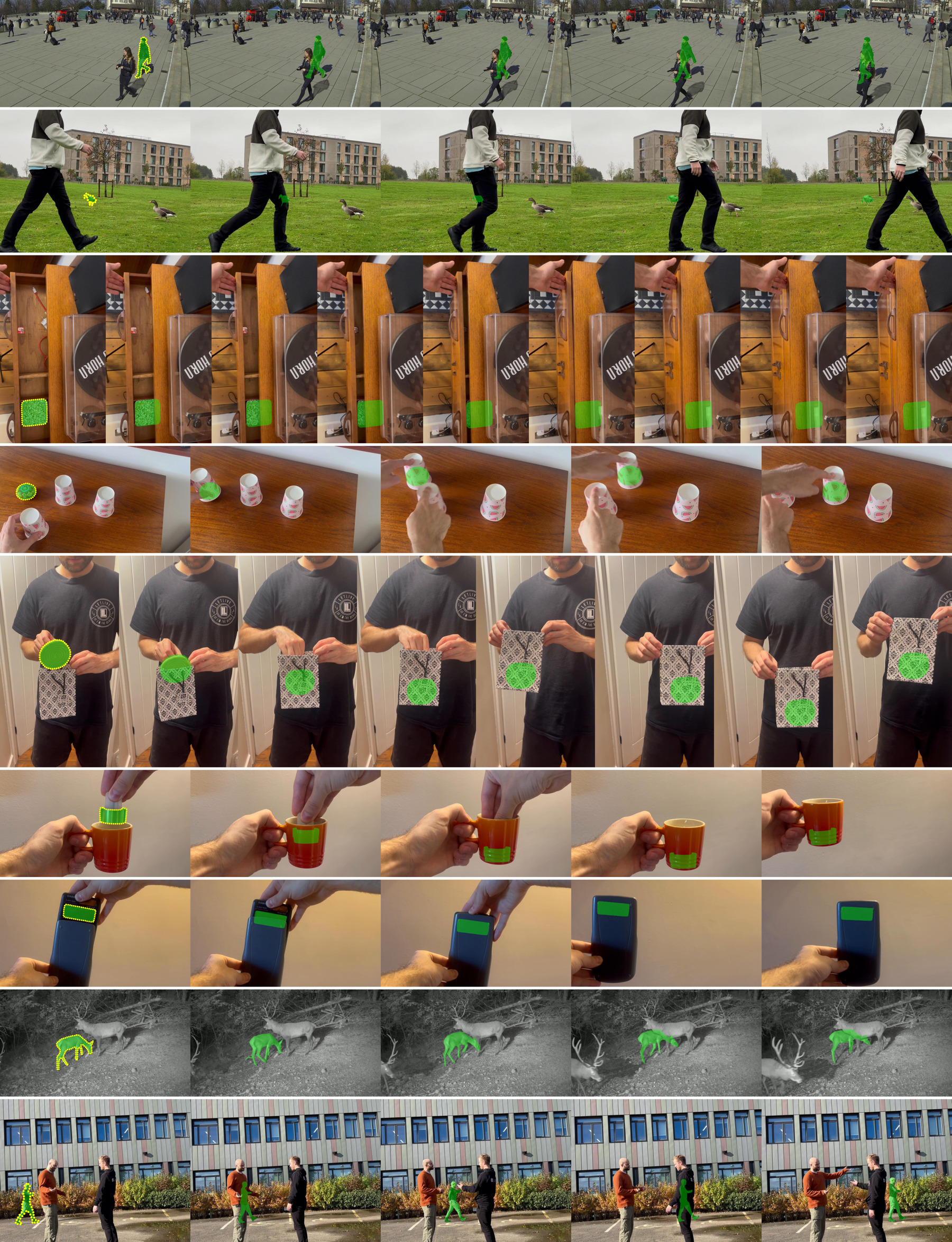}
    \caption{Further examples from the Real-VAS dataset. The yellow dashed line indicates the initial query mask, followed by the ground-truth amodal segmentation frames.}
    \label{fig:DatasetExamplesSuppl}
\end{figure}

\end{document}